\documentclass[letterpaper]{article} 
\usepackage{aaai25}  
\usepackage{times}  
\usepackage{helvet}  
\usepackage{courier}  
\usepackage[hyphens]{url}  
\usepackage{graphicx} 
\usepackage{natbib}  
\usepackage{caption} 
\usepackage{algorithm}
\usepackage{algorithmicx}
\usepackage{algpseudocode}
\usepackage{amsmath}
\usepackage{amsthm}
\usepackage[switch]{lineno}
\usepackage{amsfonts,amssymb,mathrsfs}
\usepackage{rotating}
\usepackage{multirow}
\usepackage{color, xcolor, soul}
\definecolor{pcolor}{RGB}{153,102,204}
\definecolor{mcolor}{RGB}{255,153,102}
\usepackage{longtable}
\usepackage{supertabular}
\usepackage{subfig}

\usepackage{newfloat}
\usepackage{listings}
\DeclareCaptionStyle{ruled}{labelfont=normalfont,labelsep=colon,strut=off} 
\floatstyle{ruled}
\newfloat{listing}{tb}{lst}{}
\floatname{listing}{Listing}
\nocopyright

\title{ConfigX: Modular Configuration for Evolutionary Algorithms \\via Multitask Reinforcement Learning}
\author{
    Hongshu Guo\textsuperscript{\rm 1}\equalcontrib,
    Zeyuan Ma\textsuperscript{\rm 1}\equalcontrib,
    Jiacheng Chen\textsuperscript{\rm 1},
    Yining Ma\textsuperscript{\rm 2},\\
    Zhiguang Cao\textsuperscript{\rm 3},
    Xinglin Zhang\textsuperscript{\rm 1},
    Yue-Jiao Gong\textsuperscript{\rm 1}\thanks{Corresponding author.}
}
\affiliations{
    \textsuperscript{\rm 1}South China University of Technology, 
    \textsuperscript{\rm 2}Massachusetts Institute of Technology, 
    \textsuperscript{\rm 3}Singapore Management University\\

}

\usepackage{bibentry}
\begin{document}

\maketitle

\begin{abstract}
Recent advances in Meta-learning for Black-Box Optimization (MetaBBO) have shown the potential of using neural networks to dynamically configure evolutionary algorithms~(EAs), enhancing their performance and adaptability across various BBO instances. However, they are often tailored to a specific EA, which limits their generalizability and necessitates retraining or redesigns for different EAs and optimization problems. To address this limitation, we introduce ConfigX, a new paradigm of the MetaBBO framework that is capable of learning a universal configuration agent (model) for boosting diverse EAs. To achieve so, our ConfigX first leverages a novel modularization system that enables the flexible combination of various optimization sub-modules to generate diverse EAs during training. Additionally, we propose a Transformer-based neural network to meta-learn a universal configuration policy through multitask reinforcement learning across a designed joint optimization task space. Extensive experiments verify that, our ConfigX, after large-scale pre-training, achieves robust zero-shot generalization to unseen tasks and outperforms state-of-the-art baselines. Moreover, ConfigX exhibits strong lifelong learning capabilities, allowing efficient adaptation to new tasks through fine-tuning. Our proposed ConfigX represents a significant step toward an automatic, all-purpose configuration agent for EAs.
\end{abstract}

\section{Introduction}\label{sec:intro}
Over the decades, Evolutionary Algorithms~(EAs) such as Genetic Algorithm~(GA)~\citep{GA}, Particle Swarm Optimization~(PSO)~\citep{PSO} and Differential Evolution~(DE)~\citep{DE} have been extensively researched to tackle challenging Black-Box Optimization (BBO) problems, where neither the mathematical formulation nor additional derivative information is accessible. On par with the development of EAs, one of the most crucial research avenues is the Automatic Configuration~(AC) for EAs~\citep{AC,AC-survey}. Generally speaking, AC for EAs aims at identifying the optimal configuration $c^*$ from the configuration space $\mathcal{C}$ of an evolutionary algorithm $A$, across a set of BBO problem instances $\mathcal{I}$:
\begin{equation}\label{eq:AC}
    c^* = \mathop{\arg\max}\limits_{c \in \mathcal{C}} \mathop{\mathbb{E}}\limits_{p \in \mathcal{I}}\left[ Perf(A,c,p)\right]
\end{equation}
where $Perf()$ denotes the performance of a configuration for the algorithm under a given problem instance.

\begin{figure}[t]
    \centering   
    \includegraphics[width=0.75\columnwidth]{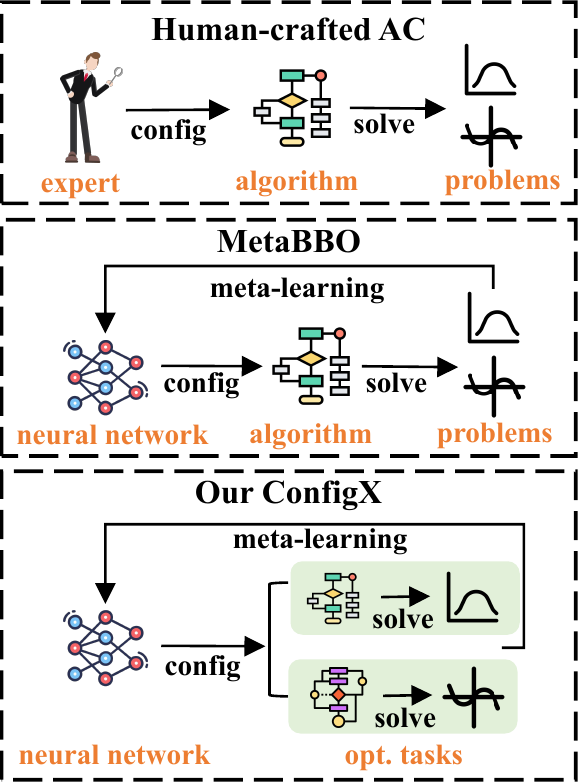}
    \caption{Conceptual overview of different AC paradigms.}
    \label{fig:overview}
\end{figure}

Traditionally, the primary research focus in AC for EAs has centered on human-crafted AC mechanisms. These mechanisms, including algorithm/operator selection~\citep{operatorselect-survey} and parameter control~\citep{parametercontrol-survey}, have demonstrated strong performance on well-known BBO benchmarks~\citep{bbob2010}, 
as well as in various eye-catching real-world scenarios such as Protein-Docking~\citep{proteinbench}, AutoML~\citep{openml}, and Prompting Optimization of Large Language Models~\citep{evoprompting}. However, a major limitation of manual AC is its heavy reliance on deep expertise. To address a specific problem, one often needs to consult experts with the necessary experience to analyze the problem and then design appropriate AC mechanisms~(as depicted in the top of Figure~\ref{fig:overview}). This impedes the broader application of EAs in diverse scientific or industrial applications. 

Recently, a novel paradigm called Meta-learning for Black-Box Optimization (MetaBBO)~\citep{metabox}, has emerged in the learning-to-optimize community. MetaBBO aims to reduce the reliance on expert-level knowledge in designing more automated AC mechanisms. As shown in the middle of Figure~\ref{fig:overview}, in MetaBBO, a neural network is meta-trained as a meta-level policy to maximize the expected performance (Eq.~(\ref{eq:AC})) of a low-level algorithm by dictating suitable configuration for solving each problem instance. By leveraging the data-driven features of deep models and the generalization strengths of meta-learning~\citep{meta-learning} across a distribution of optimization problems, these MetaBBO methods~\citep{llamoco,glhf,ribbo} have shown superior adaptability compared to traditional human-crafted AC baselines.

Despite these advancements, there remains significant potential to further reduce the expertise burden. Current MetaBBO methods often need custom neural network designs, specific learning objectives, and frequent retraining or even complete redesigns to fit different backbone EAs, overlooking the shared aspects of AC across multiple EAs. This leads to the core research question of this paper: \emph{Is it possible to develop a MetaBBO paradigm that can meta-learn an automatic, all-purpose configuration agent for diverse EAs?} We outline the detailed research objective below:
\begin{equation}\label{eq:mtac}
    c^*_k = \mathop{\arg\max}\limits_{c \in \mathcal{C}_k} \mathop{\mathbb{E}}\limits_{p \in \mathcal{I}}\left[ Perf(A_k,c,p)\right],k=1,2,...,K
\end{equation}
where $K$ is an exceedingly large number, potentially infinite. This objective is far more challenging since it can be regarded as the extension of Eq.~(\ref{eq:AC}). Concretely, it presents two key challenges: 1) \textbf{Constructing a comprehensive evolutionary algorithm space} is crucial for addressing Eq.~(\ref{eq:mtac}), from which diverse EAs can be easily sampled for meta-training the MetaBBO; 2) \textbf{Ensuring the generalization capability} of the learned meta-level policy across not only optimization problems but also various EAs is imperative.

To address these challenges, we introduce ConfigX, a pioneering MetaBBO framework capable of modularly configuring diverse EAs with a single model across different optimization problems~(as shown at the bottom of Figure~\ref{fig:overview}). 

Specifically, to address the first challenge, we present a novel modularization system for EAs, termed Modular-BBO in Section~\ref{sec:modular-bbo}. It leverages hierarchical polymorphism to efficiently encapsulate and maintain various algorithmic sub-modules within the EAs, such as mutation or crossover operators. By flexibly combining these sub-modules, Modular-BBO can generate a vast array of distinct EA structures, hence spanning a comprehensive algorithm space $\mathcal{A}$. To address the second challenge, we combine the problem instance space $\mathcal{I}$ and $\mathcal{A}$ to form a joint optimization task space $\mathcal{T}: \mathcal{A} \times \mathcal{I}$. We then consider meta-learning a Transformer based meta-level policy over moderate optimization tasks sampled from the joint space $\mathcal{T}$ to maximize the objective in Eq.~(\ref{eq:mtac}), see Section~\ref{sec:MTRL} and~\ref{sec:details}. 
For each task $T=(A_m, I_n)$, the Transformer generates configurations by conditioning on a sequence of state tokens corresponding to the sub-modules in $A_m$. 
Through large-scale multitask reinforcement learning over the sampled tasks, it yields a universal meta-policy that exhibits robust generalization to unseen algorithm structures and problem instances.

\noindent We summarize our contributions in this paper in three folds:
\begin{itemize}
    \item We introduce ConfigX, the first MetaBBO framework to learn a pre-trained universal AC agent via multitask reinforcement learning, enabling modular configuration of diverse EAs across various optimization problems.
    \item Technically, we present Modular-BBO as a novel system for EA modularization that simplifies the management of sub-modules and facilitates the sampling of diverse algorithm structures. We then propose a Transformer-based architecture to meta-learn a universal configuration policy over our defined joint optimization task space.
    \item Extensive benchmark results show that the configuration policy pre-trained by ConfigX not only achieves superior zero-shot performance against the state-of-the-art AC software SMAC3, but also exhibits favorable lifelong learning capability via efficient fine-tuning.
\end{itemize}

\begin{figure*}[t]
    \centering   
    \includegraphics[width=0.90\textwidth]{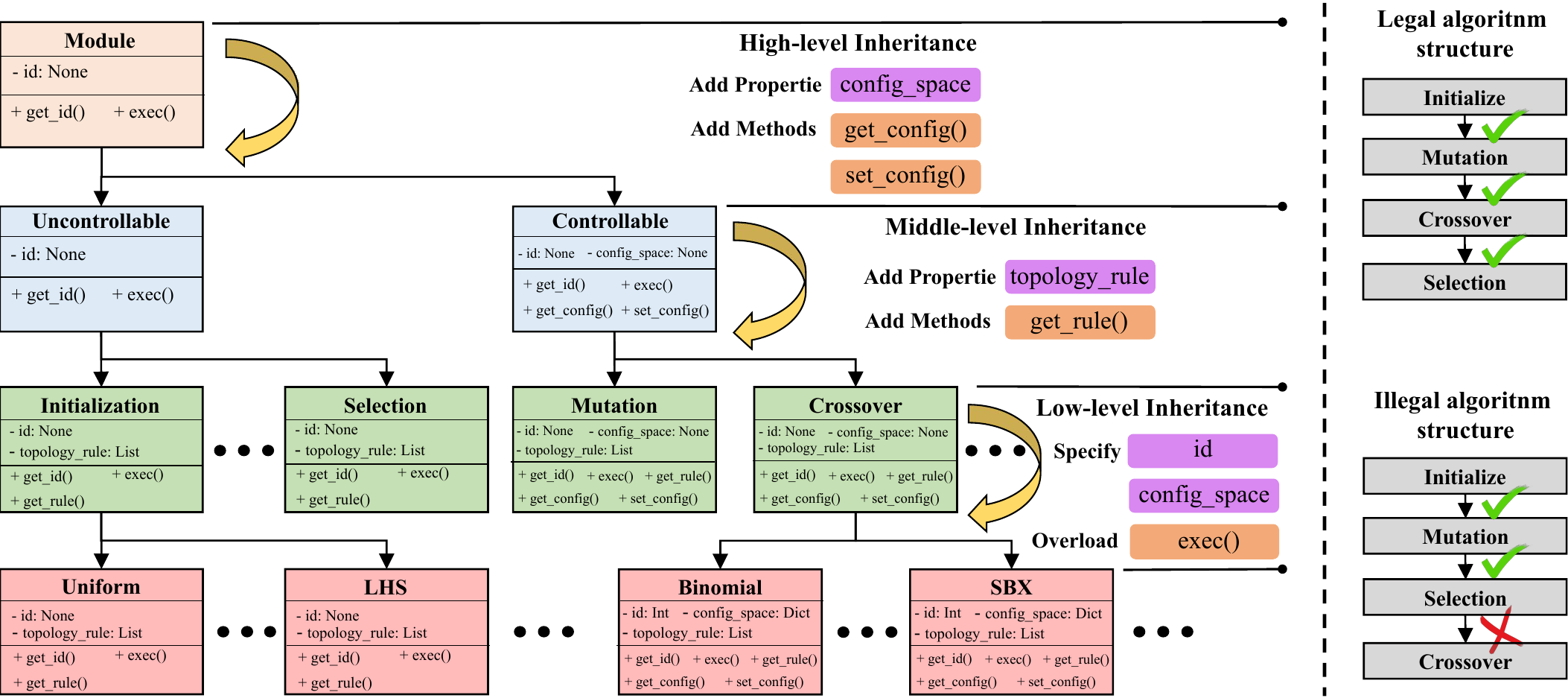}
    \caption{\textbf{Left}: The hierarchical polymorphism in Modular-BBO. \textbf{Right}: Legal/Illegal algorithm examples in Modular-BBO.}
    \label{fig:hierarchical}
\end{figure*}
\section{Related Works}
\subsection{Human-crafted AC}
Human-crafted AC mechanisms enhance the optimization robustness of EAs through two main paradigms: Operator Selection~(OS) and Parameter Control~(PC). OS is geared towards selecting proper evolutionary operators~(i.e., mutation in DE~\citep{SaDE}) for EAs to solve target optimization problems. To this end, such AC mechanism requires preparing a group of candidate operators with diverse searching behaviours. Besides, throughout the optimization progress, OS facilitates dynamic selection over the candidate operators, either by a roulette wheel upon the historical success rates~\citep{EPSO} or random replacement upon the immediate performance improvement~\citep{EPSDE}. PC, on the other hand, aims at configuring (hyper-) parameters for the operators in EAs, (e.g. the inertia weights in PSO~\cite{FEPSO} and the scale factors in DE~\cite{JADE,SHADE}), while embracing similar adaptive mechanisms as OS to achieve dynamic configuration. We note that OS and PC are complementary rather than conflicting. Recent outperforming EAs such as MadDE~\cite{MadDE}, AMCDE~\cite{AMCDE} and SAHLPSO~\cite{SAHLPSO} integrate both to obtain maximal performance gain. However, the construction of the candidate operators pool, the parameter value range in PC, and the adaptive mechanism in both of them heavily rely on expertise. A more versatile and efficient alternative for human-crafted AC is Bayesian Optimization (BO)~\cite{BO2016}. By iteratively updating and sampling from a posterior distribution over the algorithm configuration space, a recent open-source BO software SMAC3~\cite{SMAC3} achieves the state-of-the-art AC performance on many realistic scenarios.
\algdef{SE}[DOWHILE]{Do}{doWhile}{\algorithmicdo}[1]{\algorithmicwhile\ #1}
\begin{algorithm}[tb]
\caption{Algorithm Structure Generation.}
\label{alg:algorithm}\small
\textbf{Input}: All accessible modules $\mathbb{M}$, all Initialization modules $\mathbb{M}_\text{init}$ \\
\textbf{Output}: A legal algorithm structure $A$.
\begin{algorithmic}[1] 
\State Create an empty structure $A = \emptyset$, set index $j=0$
\State Randomly select an Initialization module from $\mathbb{M}_\text{init}$ as $a_j$   
\State $A \leftarrow A \bigcup a_j$

\While {not \textsc{Completed}}
\State $j=j+1$
\While {\textsc{Violated}}
\State Randomly select a module from $\mathbb{M}\backslash\mathbb{M}_\text{init}$ as $a_j$
\State Check the violation between $a_j$ and $a_{j-1}$
\EndWhile
\State $A \leftarrow A \bigcup a_j$
\EndWhile
\end{algorithmic}
\end{algorithm}

\subsection{MetaBBO}
To relieve the expertise dependency of human-crafted AC, recent MetaBBO works introduce neural network-based control policy~(typically denoted as the meta-level policy $\pi_\theta$) to automatically dictate desired configuration for EAs~\cite{ma2024toward,yang4956956meta}. Generally speaking, the workflow of MetaBBO follows a bi-level optimization process: 1) At the meta level, the policy $\pi_\theta$ configures the low-level EA and assesses its performance, termed meta performance. The policy leverages this observed meta performance to refine its decision-making process, training itself through the maximization of accumulated meta performance, thereby advancing its meta objective. 2) At the lower level, the BBO algorithm receives a designated algorithmic configuration from the meta policy. With this configuration in hand, the low-level algorithm embarks on the task of optimizing the target objective. It observes the changes in the objective values and relays this information back to the meta optimizer, contributing to the meta performance signal. Similarly, existing MetaBBO works focus predominantly on OS and PC. Although a predefined operator group remains necessary, the selection decisions in works on OS~\cite{DEDDQN,DEDQN,RLEMMO} are made by the meta policy $\pi_\theta$ which relieves the expert-level knowledge requirement. A notable example is RL-DAS~\cite{RLDAS} where advanced DE algorithms are switched entirely for complementary performance. In PC scenarios, initial works parameterize $\pi$ with simple Multi-Layer Perceptron  (MLP)~\cite{RLPSO,RLHPSDE} and Long Short-Term Memory (LSTM)~\cite{LDE}, whereas the latest work GLEET~\cite{GLEET} employs Transformer~\cite{transformer} architecture aiming at more versatile configuration. Besides, works jointly configure both OS and PC such as MADAC~\cite{MADAC} also show robust performance on complex problems~\cite{dacbench}.

\section{Methodology}\label{sec:method}
In this section, we elaborate on ConfigX step by step. We first explain in Section~\ref{sec:modular-bbo} the design of Modular-BBO and how to use it for efficient generation of diverse algorithm structures. We next provide a Markov Decision Process~(MDP) definition of an optimization task and derive the corresponding multi-task learning objective in Section~\ref{sec:MTRL}. At last, we introduce in Section~\ref{sec:details} the details of each component in the defined MDP and the proposed Transformer based architecture.
\subsection{Modular-BBO}\label{sec:modular-bbo}
As illustrated in the left of Figure~\ref{fig:hierarchical}, the design philosophy of Modular-BBO adheres to a Hierarchical Polymorphism in \emph{Python} which ensures the ease of maintaining different sub-modules~(third-level sub-classes in Figure~\ref{fig:hierarchical}, labeled in green), as well as their practical variants~(bottom-level sub-classes in Figure~\ref{fig:hierarchical}, labeled in red) in modern EAs. By facilitating the high-to-low level inheritances, Modular-BBO provides universal programming interfaces for the modularization of EAs, along with essential module-specific properties/methods to support diverse behaviours of various sub-modules. Further elaboration on each inheritance level is provided below.
\paragraph{High-level.}
All sub-module classes in Modular-BBO stem from an abstract base class \textsc{Module}. It declares universal properties/interfaces shared among various sub-module variants, yet leave them void. At high-level inheritance, two sub-classes \textsc{Uncontrollable} and \textsc{Controllable} inherit from \textsc{Module}. The two sub-classes divide all possible sub-modules in modern EAs into the ones with (hyper-) parameters and those without. For \textsc{Controllable} modules, we declare its (hyper-) parameters by adding a 
\textit{config\_space} 
property. Additionally, we include the corresponding 
\textit{get\_config}() and \textit{set\_config}() 
methods for configuring the (hyper-) parameters. Currently, these properties and methods remain void until a specific EA sub-module is instantiated.

\paragraph{Middle-level.} At this inheritance level, \textsc{Uncontrollable} and \textsc{Controllable} are further divided into common sub-modules in EAs, e.g., initialization, mutation, selection and etc. 
To combine these sub-modules legally and generate legal algorithm structures, 
we introduce module-specific 
\textit{topology\_rule} 
as a guidance during the generating process~(Algorithm~\ref{alg:algorithm}), by invoking the added 
\textit{get\_rule}() 
method. We present a pair of examples in the left of Figure~\ref{fig:hierarchical} to showcase one of the possible violation during the algorithm structure generation, where \textsc{Crossover} is not allowed after \textsc{Selection} is a common sense in EAs.

\paragraph{Low-level.} Within the low-level inheritance, we borrow from a large body of EA literature diverse practical sub-module variants~(i.e., lots of initialization strategy have been proposed in literature such as Sobol sampling~\citep{sobol} and LHS sampling~\citep{LHS}) and maintaining them by inheriting from the sub-module classes in middle-level inheritance. When inheriting from the parent class, a concrete sub-module variant has to instantiate void modules
\textit{id} and \textit{config\_space}, which detail its unique identifier in Modular-BBO and controllable parameters respectively. It also have to overload 
\textit{exec}() 
method by which it operates the solution population. The unique module id of a sub-module variant is a $16$-bit binary code of which: 1) the first bit is $0$ or $1$ to denote if this variant is \textsc{Uncontrollable} or \textsc{Controllable}. 2) the $2$-nd to $7$-th bits denote the sub-module category~(third-level sub-classes in Figure~\ref{fig:hierarchical}, labeled in green) to which the variant belongs. 3) the last $9$ bits denotes its id within this sub-module category. 

For now, Modular-BBO has included $11$ common sub-module categories in EAs: \textsc{Initialization}~\cite{initsurvey}, \textsc{Mutation}~\cite{desurvey}, \textsc{Crossover}~\cite{spears1995adapting}, \textsc{PSO\_Update}~\cite{psosurvey}, \textsc{Boundary\_Control}~\cite{kadavy2023impact}, \textsc{Selection}~\cite{ga_selectsurvey}, \textsc{Multi\_Strategy}~\cite{opselect_survey}, \textsc{Niching}~\cite{multipop}, \textsc{Information\_Sharing}~\cite{toulouse1996communication}, \textsc{Restart\_Strategy}~\cite{restartsurvey}, \textsc{Population\_Reduction}~\cite{popsize_survey}. We construct a collection of over $100$ variants for these sub-module categories from a large body of literature and denote this collection as module space $\mathbb{M}$. Theoretically, by using the algorithm generation process described in Algorithm~\ref{alg:algorithm}, Modular-BBO spans a massive algorithm structure space $\mathcal{A}$ containing millions of algorithm structures. Due to the limitation of space, we provide the detail of each sub-module variant in $\mathbb{M}$ in Appendix A, 
Table 1, including the id, name, type, configuration space, topology rule and functional description. We also provide a detailed explanation for Algorithm~\ref{alg:algorithm} in Appendix B. 

\begin{figure*}[t]
    \centering   
    \includegraphics[width=0.85\textwidth]{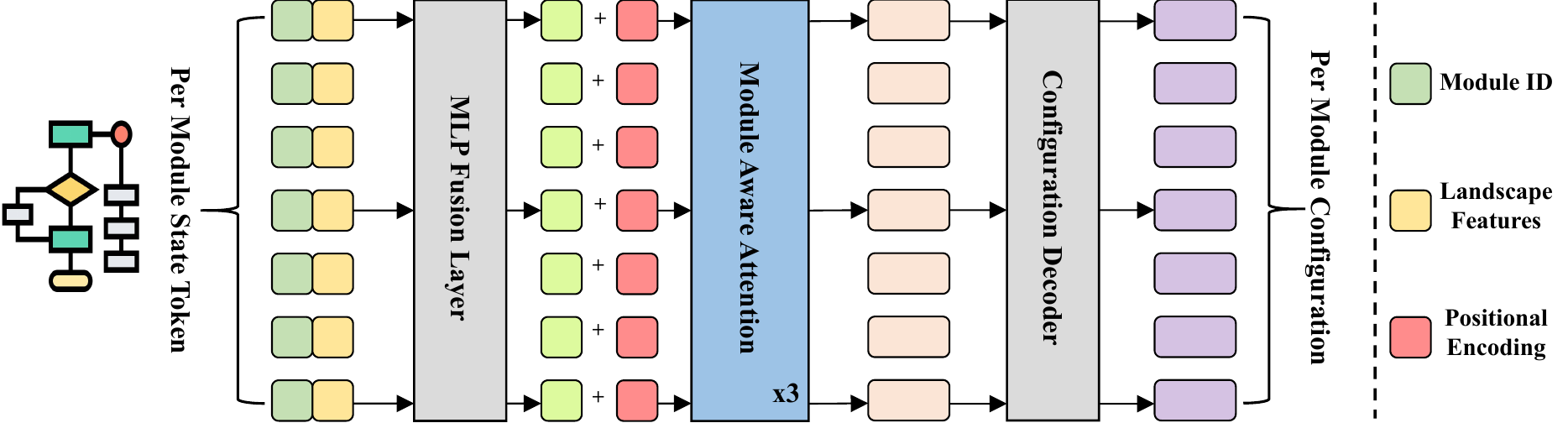}
    \caption{The workflow of the Transformer based configuration policy in ConfigX. }
    \label{fig:nn}
\end{figure*}
\subsection{Multi-task Learning in ConfigX}\label{sec:MTRL}
\subsubsection{Optimization Task Space} We first define an optimization task space $\mathcal{T}$ as a synergy of an algorithm space $\mathcal{A}$ and an optimization problem set $\mathcal{I}$. Then an optimization task $T \in \mathcal{T}$ can be defined as $T:\{A\in\mathcal{A}, p \in \mathcal{I}\}$. In this paper, we adopt the algorithm space spanned by Modular-BBO as $\mathcal{A}$, and the problem instances from well-known CoCo-BBOB benchmark~\citep{bbob2010}, Protein-docking benchmark~\cite{proteinbench} and HPO-B benchmark~\cite{hpo-b} as $\mathcal{I}$.
\subsubsection{AC as an MDP} For an optimization task $T:\{A, p\}$, we facilitate a Transformer based policy $\pi_\theta$~(detailed in Section~\ref{sec:details}) to dynamically dictate desired configuration for $A$ to solve $p$. This configuration process can be formulated as an Markov Decision Process~(MDP) $\mathcal{M} := (S, C, \Gamma, R, H, \gamma)$, where $S$ denotes the state space that reflect optimization status, $C$ denotes the action space which is exactly the configuration space of algorithm $A$, $\Gamma(s_{t+1}|s_t,c_t)$ denotes the optimization transition dynamics, $R(s_t, c_t)$ measures the single step optimization improvement obtained by using configuration $c_t$ for optimizing $p$. $H$ and $\gamma$ are the number of optimization iterations and discount factor respectively. At each optimization step $t$, the policy $\pi_\theta$ receives a state $s_t$ and then outputs a configuration $c_t=\pi_\theta(s_t)$ for $A$. Using $c_t$, algorithm $A$ optimizes the optimization problem $p$ for a single step. The goal is to find an optimal policy $\pi_{\theta^*}$ which maximizes the accumulated optimization improvement during the optimization process: $G = \sum_{t=1}^H\gamma^{t-1}R(s_t,c_t)$. Recall that our ConfigX aims at addressing a more challenging objective in Eq.~(\ref{eq:mtac}), where the goal is to maximize the accumulated optimization improvement $G$ of all tasks $T\in\mathcal{T}$. We use $s_t^i$ and $c_t^i$ to denote the input state and the outputted configuration of the policy $\pi_\theta$ for the $i$-th task in $\mathcal{T}$. Then the objective in Eq.~(\ref{eq:mtac}) can be rewritten as a multi-task RL problem:
\begin{equation}\label{ppo-obj}
    \mathbb{J}(\theta) =\frac{1}{K\cdot N} \sum_{i=1}^{K\cdot N}\sum_{t=1}^H\gamma^{t-1}R(s_t^i,c_t^i)
\end{equation}
where we sample $K\cdot N$ tasks from $\mathcal{T}$ to train $\pi_\theta$ since the number of tasks in $\mathcal{T}$ is massive. These tasks is sampled first by calling Algorithm~\ref{alg:algorithm} $K$ times to obtain $K$ algorithm structures, and then combine these algorithm with the $N$ problem instances in $\mathcal{I}$. In this paper we use Proximal Policy Optimization~(PPO)~\citep{ppo}, a popular policy gradient~\citep{policygradient} method for optimizing this objective in a joint policy optimization~\citep{metamorph} fashion. We include the pseudocode of the RL training in Appendix D. 

\subsection{ConfigX}\label{sec:details}
Progress in MetaBBO has made it possible to meta-learn neural network-based control policies for configuring the backbone EAs to solve optimization problems. However, existing MetaBBO methods are not suitable for the massive algorithm structure space $\mathcal{A}$ spanned by the module space $\mathbb{M}$ in our proposed Modular-BBO, since learning a separate policy for each algorithm structure is impractical. However, the modular nature of EAs implies that while each structure is unique, they are still constructed from the same module space and potentially shares sub-modules and workflows with other algorithm structures. We now describe how ConfigX exploits this insight to address the challenge of learning a universal controller for different algorithm structures.

\subsubsection{State Design}\label{sec:state}
In ConfigX, we encode not only the algorithm structure information but also the optimization status information into the state representation to ensure the generalization across optimization tasks. Concretely, as illustrated in the left of Figure~\ref{fig:nn}, for $i$-th tasks $T_i:\{A_i,p_i\}$ in the sampled $K \cdot N$ tasks, we encode a information pair for each sub-module in $A_i$, e.g., $s_i:\{s_{i,j}^\text{id}, s_i^\text{opt}\}_{j=1}^{L_i}$, where $s_{i,j}^\text{id} \in \{0,1\}^{16}$ denotes the unique module id for $j$-th sub-module in $A_i$, $s_i^\text{opt} \in \mathbb{R}^9$ denotes the algorithm performance information which we borrow the idea from recent MetaBBO methods~\citep{RLDAS,neurela}, $L_i$ denotes the number of sub-modules in $A_i$. We provide details of these information pairs in Appendix C. 

\subsubsection{State Encode}
We apply an MLP fusion layer to preprocess the state representation $s_i$. This fusion process ensures the information within the information pair $\{s_{i,j}^\text{id}, s_i^\text{opt}\}$ join each other smoothly~(as illustrated in left part of Figure~\ref{fig:nn}).
\begin{equation}\label{eq:idemb}
\begin{split}
    \hat{e}_{i,j} = &\text{hstack}(\phi(s_{i,j}^\text{id}; \textbf{W}_{e}^\text{id}); \phi(s_i^\text{opt}; \textbf{W}_{e}^\text{opt})) \\
    e_{i,j} = &\phi(\hat{e}_{i,j}; \textbf{W}_{e}),\quad j=1,...,L_i
\end{split}
\end{equation}
Where $\phi(\cdot;\textbf{W}_{e}^\text{id})$, $\phi(\cdot;\textbf{W}_{e}^\text{opt})$ and $\phi(\cdot;\textbf{W}_{e})$ denotes MLP layers with the shape of $16\times16$, $9\times 16$ and $32 \times 64$ respectively, $e_{i,j}$ denotes the fused information for each sub-module. Then we add $Sin$ Positional Encoding~\cite{transformer} $\textbf{W}_\text{pos}$ to each sub-module, which represents the relative position information among all sub-modules in the algorithm structure.
\begin{equation}
    \textbf{h}_{i}^{(0)} = \text{vstack}(e_{i,j};\cdots;e_{i,L_i}) + \textbf{W}_\text{pos}
\end{equation}
where $\textbf{h}_{i}^{(0)} \in \mathbb{R}^{L_\text{max} \times 64}$ denotes the module embedding for each sub-module. We note that since the number of sub-modules~($L_i$) may vary between different algorithm structures, we zero pad $\textbf{h}_{i}^{(0)}$ to a pre-defined maximum length $L_\text{max}$ to ensure input size invariant among tasks. 

\subsubsection{Module Aware Attention}
From the module embeddings $\textbf{h}_{i}^{(0)}$ described above, we obtains the output features for all sub-modules as: 
\begin{equation}
\begin{split}
    \hat{\textbf{h}}^{(l)}_i = & \text{LN}(\text{MSA}(\textbf{h}^{(l-1)}_i) + \textbf{h}^{(l-1)}_i), l=1,2,3\\
    \textbf{h}^{(l)}_i = & \text{LN}(\phi(\hat{\textbf{h}}_i; \textbf{W}^{(l)}_F)) + \hat{\textbf{h}}_i^{(l)}), l=1,2,3
\end{split}
\end{equation}
where LN is Layernorm~\cite{ba2016layer}, MSA is Multi-head Self-Attention~\cite{transformer} and $\phi(\cdot;\textbf{W}^{(l)}_F)$ are MLP layers with the shape of $64\times64$. In this paper we use $l=3$ MSA blocks to process the module embeddings~(as illustrated in the middle of Figure~\ref{fig:nn}).

\begin{figure*}[t]
    \centering   
    \includegraphics[width=0.9\textwidth]{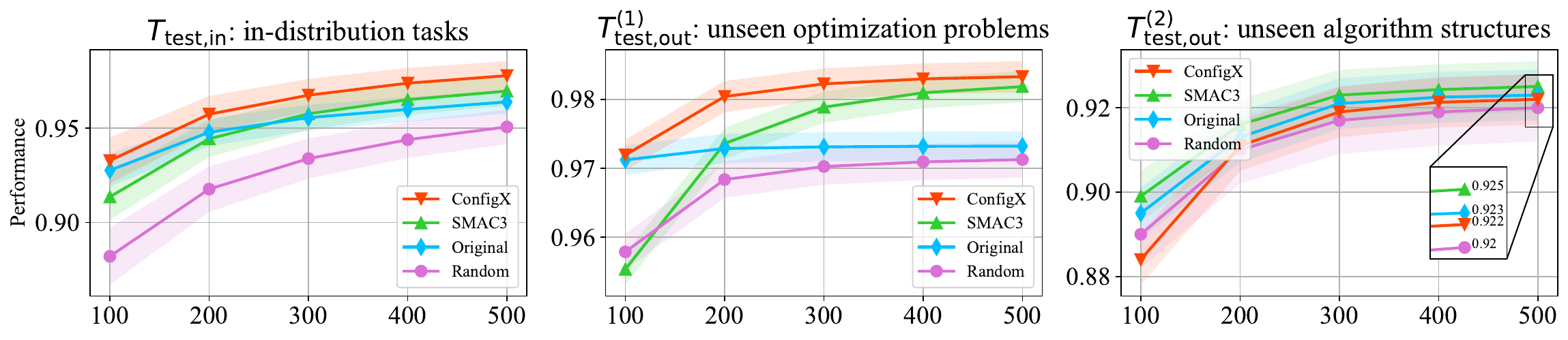}
    \caption{Optimization curves of the pre-trained ConfigX model and the baselines, over three different zero-shot scenarios.}
    \label{fig:rand_de_hist}
\end{figure*}
\subsubsection{Configuration Decoder}\label{sec:action}
In ConfigX, the policy $\pi_\theta(c_i|s_i)$ models the conditional distribution of $A_i$'s configuration $c_i$ given the state $s_i$. As illustrated in the right of Figure~\ref{fig:nn}, for each sub-module $a_j$ in an algorithm $A_i=\{a_1, a_2,....\}$, we output distribution parameters $\mu$ and $\Sigma$ as:
\begin{equation}
\begin{split}
    \mu_j = \phi(h_{i,j}^{(3)}; \textbf{W}_\mu), \quad&\Sigma_j = \text{Diag} \phi(h_{i,j}^{(3)}; \textbf{W}_\Sigma) \\
    c_{i,j} \sim & \mathcal{N}(\mu_j; \Sigma_j)
\end{split}
\end{equation}
where $\phi(\cdot;\textbf{W}_\mu)$ and $\phi(\cdot;\textbf{W}_\Sigma)$ are two MLP layers with the same shape of $64\times C_\text{max}$, $c_{i,j} \in \mathbb{R}^{C_\text{max}}$ denotes the configurations for sub-module $a_j$ in algorithm structure $A_i$. 
Since the size of the configuration spaces may vary between different sub-modules, we pre-defined a maximum configuration space size $C_\text{max}$ to cover the sizes of all sub-modules. If the size of a sub-module is less than $C_\text{max}$, we use the first few configurations in $c_{i,j}$ and ignore the rest. 

 For the critic, we calculate the value of a sub-module as $V(s_{i,j}) = \phi(\textbf{h}_{i,j}^{(3)}; \textbf{W}_c)$ using a MLP with the shape of $\textbf{W}_c\in\mathbb{R}^{64\times16\times1}$. The value of the algorithm structure is the averaged value per sub-module $V(s_{i}) = \frac{1}{L_\text{max}}\sum_{j=1}^{L_\text{max}} V(s_{i,j})$.

\subsubsection{Reward Function}\label{sec:reward}

The objective value scales across different problem instances can vary. To ensure the accumulated performance improvement across tasks approximately share the same numerical level, we propose a task agnostic reward function. At optimization step $t$, the reward function on any problem instance $p\in\mathcal{I}$ is formulated as:
\begin{equation}
    r_t = \delta\times\frac{f^*_{p,t-1} - f^*_{p,t}}{f^*_{p,0} - f_p^*}
\end{equation}
where $f^*_{p,t}$ is the found best objective value of problem instance $p$ at step $t$, $f_p^*$ is the global optimal objective value of $p$ and $\delta=10$ is a scale factor. In this way, we make the scales of the accumulated improvement in all tasks similar and hence stabilize the training.

\section{Experiment}

In this section, we discuss the following research questions: 
\textbf{RQ1}: Can pre-trained ConfigX model zero-shots to unseen tasks with unseen algorithm structures and/or unseen problem instances?
\textbf{RQ2}: If the zero-shot performance is not as expected, is it possible to fine-tune ConfigX to address novel algorithm structures in future?
\textbf{RQ3}: How do the concrete designs in ConfigX contribute to the learning effectiveness? 
Below, we first introduce the experimental settings and then address RQ1$\sim$RQ3 respectively. 

\subsection{Experimental Setup}

\subsubsection{Training setup.} 
We have prepared several task sets from different sub-task-spaces of the overall task space $\mathcal{T}$~(defined at Section~\ref{sec:MTRL}) to aid for the following experimental validation. Concretely, denote $\mathcal{I}_\text{syn}$ as the problems in CoCo-BBOB suite, $\mathcal{I}_\text{real}$ as all realistic problems in Protein-docking benchmark and HPO-B benchmark, $\mathcal{A}_\text{DE}$ as the algorithm structure space only including DE variants, $\mathcal{A}_\text{PSO,GA}$ as the algorithm structure space including PSO and GA variants, we have prepared $256$ optimization tasks as training task set $T_\text{train} \subset \mathcal{A}_\text{DE} \times \mathcal{I}_\text{syn}$ , another $512$ optimization tasks as in-distribution testing task set $T_\text{test,in} \subset \mathcal{A}_\text{DE} \times \mathcal{I}_\text{syn}$. For out-of-distribution tasks, we have prepared two task sets: $T_\text{test,out}^{(1)} \subset \mathcal{A}_\text{DE} \times \mathcal{I}_\text{real}$ and $T_\text{test,out}^{(2)} \subset \mathcal{A}_\text{PSO,GA} \times \mathcal{I}_\text{syn}$, each with $512$ task instances. During the training, for a batch of $batch\_size=32$ tasks, PPO~\citep{ppo} method is used to update the policy net and critic $\kappa=3$ times for every $10$ rollout optimization steps. All of the tasks are allowed to be optimized for $H=500$ optimization steps. The training lasts for $50$ epochs with a fixed learning rate $0.001$. All experiments are run on an Intel(R) Xeon(R) 6348 CPU with 504G RAM. Refer to Appendix E.1 
for more details.

\subsubsection{Baselines and Performance Metric.} In the following comparisons, we consider three baselines: \textbf{SMAC3}, which is the state-of-the-art AC software based on Bayesian Optimization and aggressive racing mechanism; 
\textbf{Original}, which denotes using the suggested configurations in sub-modules' original paper (see Appendix A 
for one-to-one correspondence); \textbf{Random}, which randomly sample the configurations for the algorithm from the algorithm's configuration space. For the pre-trained model in ConfigX and the above baselines, we calculate the performance of them on tested task set by applying them to configure each tested task for $51$ independent runs and then aggregate a normalized accumulated optimization improvement across all tasks and all runs, we provide more detailed calculation steps in Appendix E.2. 

\subsection{Zero-shot Performance (RQ1)}\label{sec:zeroshot}

We validate the zero-shot performance of ConfigX by first pre-training a model on $T_\text{train}$. Then the pre-trained model is directly used to facilitate AC process for tasks in tested set, without any fine-tuning. Concretely, we aims at validating the zero-shot generalization performance in three different scenarios: 1) $T_\text{test,in}$, where the optimization tasks come from the same task space on which ConfigX is pre-trained. 2) $T_\text{test,out}^{(1)}$, where the optimization tasks locate beyond the optimization problem scope of the training task space. 3) $T_\text{test,out}^{(2)}$, where the optimization tasks locate beyond the algorithm structure scope of the training task space. We present the optimization curves of our pre-trained model and the baselines in Figure~\ref{fig:rand_de_hist}, where the x-axis denotes the optimization horizon and y-axis denotes the performance metric we defined previously. The results in Figure~\ref{fig:rand_de_hist} reveal several key observations: 1) In all zero-shot scenarios, ConfigX presents significantly superior performance to the Random baseline, which randomly configures the algorithms in the tested tasks. This underscores the effectiveness of the multi-task reinforcement learning in ConfigX. 2) The results on $T_\text{test,in}$ demonstrate that pre-training ConfigX on some task samples of the given task space is enough to ensure the generalization to the other tasks within this space, surpassing the state-of-the-art AC baseline SMAC3. 3) The results on $T_\text{test,out}^{(1)}$ show that ConfigX is capable of adapting itself to totally unseen optimization problem scope. This observation attributes to our state representation design, where the optimization status borrowed from recent MetaBBO works are claimed to be generic across different problem scopes. 4) Though promising, we find that the zero-shot performance of ConfigX on $T_\text{test,out}^{(2)}$ is not as expected. It is not surprising as the sub-modules and structures in GA/PSO are significantly different from those in DE, which hinders ConfigX from applying its configuration experience with DE on PSO/GA 
optimization tasks. We explore whether this generalization gap could be addressed through further fine-tuning in the next section.   
\begin{figure}[t]
    \centering   
    \includegraphics[width=0.99\columnwidth]{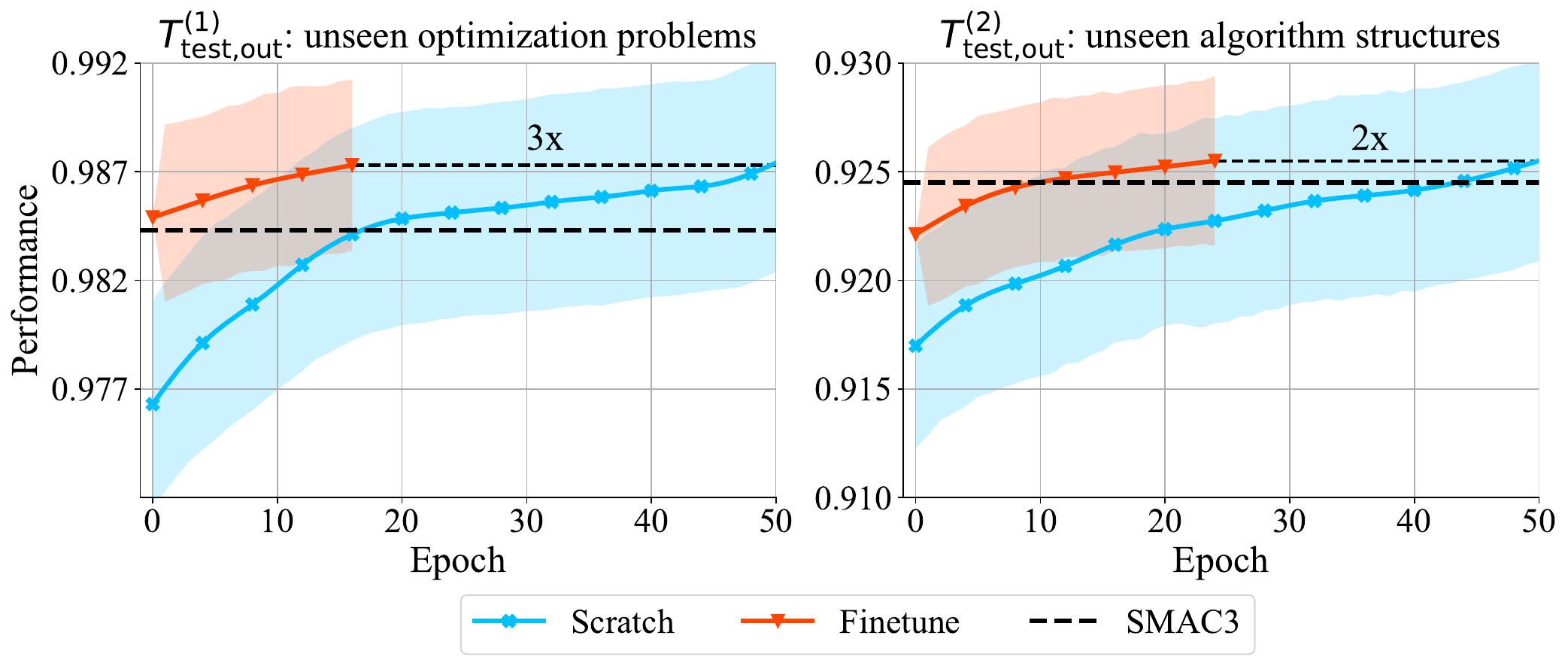}
    \caption{The learning curves of fine-tuning and re-training ConfigX on novel optimization problems or algorithm structures. The fine-tuning saves $3$x and $2$x learning steps than the re-training on $T_\text{test,out}^{(1)}$ and $T_\text{test,out}^{(2)}$ respectively.}
    \label{fig:lll}
\end{figure}
\subsection{Lifelong Learning in ConfigX (RQ2)}
The booming algorithm designs in EAs, together with the increasingly diverse optimization problems pose non-negligible challenges to universal algorithm configuration methods such as our ConfigX. On the one hand, although our pre-trained model shows uncommon AC performance when encountered with novel optimization problems~(middle of Figure~\ref{fig:rand_de_hist}), further performance boost is still needed especially in industrial scenarios. On the other hand, as shown in the left of Figure~\ref{fig:rand_de_hist}, the pre-trained model can not cover those algorithm sub-modules which have not been included within its training algorithm structure space. Both situations above underline the importance of lifelong learning in ConfigX. We hence investigate the fine-tuning efficiency of the pre-trained model in this section. Concretely, we compare the learning curves of 1) fine-tuning the pre-trained model, and 2) re-training a new model from scratch in Figure~\ref{fig:lll}, where the x-axis denotes the learning epochs and the y-axis denotes the aforementioned performance metric over the tested task set. The results reveal that ConfigX supports efficient fine-tuning for adapting out-of-distribution optimization tasks, which in turn provides operable guidance for lifelong learning in ConfigX: (a) One can configure an algorithm already included in the algorithm structure space of our Modular-BBO, yet on different problem scope, by directly using the pre-trained model. (b) One can also integrate novel algorithm designs into Modular-BBO and then facilitate efficient fine-tuning to enhance the performance of the pre-trained model on these novel algorithm structures.

\subsection{Ablation Study (RQ3)}

\begin{table}[t]
\centering
\resizebox{0.85\columnwidth}{!}{%
\begin{tabular}{c|cccc}
\hline
 &
  $T_\text{test,in}$ &
  $T^{(1)}_\text{test,out}$ &
  $T^{(2)}_\text{test,out}$ &
  \\ \hline
ConfigX     
&  \begin{tabular}[c]{@{}c@{}}9.81E-01\\ $\pm$7.33E-03\end{tabular}
&  \textbf{\begin{tabular}[c]{@{}c@{}}9.86E-01\\ $\pm$2.64E-03\end{tabular}}
&  \textbf{\begin{tabular}[c]{@{}c@{}}9.22E-01\\ $\pm$6.94E-03\end{tabular}}
\\  \hline
ConfigX-MLP 
&  \begin{tabular}[c]{@{}c@{}}9.70E-01\\ $\pm$8.13E-03\end{tabular}
&  \begin{tabular}[c]{@{}c@{}}9.80E-01\\ $\pm$2.54E-03\end{tabular}
&  \begin{tabular}[c]{@{}c@{}}9.16E-01\\ $\pm$6.57E-03\end{tabular}
\\ \hline
ConfigX-LPE 
&  \textbf{\begin{tabular}[c]{@{}c@{}}9.82E-01\\ $\pm$7.62E-03\end{tabular}}
&  \begin{tabular}[c]{@{}c@{}}9.84E-01\\ $\pm$2.58E-03\end{tabular}
&  \begin{tabular}[c]{@{}c@{}}9.20E-01\\ $\pm$6.89E-03\end{tabular}
\\ \hline
ConfigX-NPE 
&  \begin{tabular}[c]{@{}c@{}}9.74E-01\\ $\pm$7.75E-03\end{tabular}
&  \begin{tabular}[c]{@{}c@{}}9.81E-01\\ $\pm$2.67E-03\end{tabular}
&  \begin{tabular}[c]{@{}c@{}}9.19E-01\\ $\pm$6.73E-03\end{tabular}
\\ \hline
MLP-NPE     
&  \begin{tabular}[c]{@{}c@{}}9.51E-01\\ $\pm$9.27E-03\end{tabular}
&  \begin{tabular}[c]{@{}c@{}}9.73E-01\\ $\pm$2.71E-03\end{tabular}
&  \begin{tabular}[c]{@{}c@{}}9.06E-01\\ $\pm$7.29E-03\end{tabular}
\\ \hline
\end{tabular}%
}
\caption{Performance of different ablated baselines.}
\label{tab:abla}
\end{table}

In Section~\ref{sec:details}, we proposed a Transformer based architecture to encode and process the state information of all sub-modules within an algorithm structure. In particular, we added $Sin$ positional embeddings~(PE) to each sub-module token as additional topology structure information for learning. We further apply Multi-head Self-Attention~(MSA) to enhance the module-aware information sharing. In this section we investigate on what extent these designs influence ConfigX's learning effectiveness. Concretely, for the positional embeddings, we introduce two ablations 1) ConfigX-NPE: remove the $Sin$ PE from ConfigX. 2) ConfigX-LPE, replace the $Sin$ PE by Learnable PE~\citep{learntPE}. For the MSA, we introduce one ablation ConfigX-MLP: cancel the information sharing between the sub-modules by replacing the MSA blocks by an MLP layer. We present the final performance of these baselines and ConfigX on the tested task sets in Table~\ref{tab:abla}. The results underscores the importance of these special designs: (a) Without the MSA block, ConfigX struggles in learning the configuration policy in an informative way. (b) Without the positional embdeddings, the configuration policy in ConfigX becomes agnostic to the structure information of the controlled algorithm. (c) Learnable PE shows similar performance with $Sin$ PE, while introducing additional parameters for ConfigX to learn.

\section{Conclusion}

In this paper, we propose ConfigX as a pioneer research exploring the possibility of learning a universal MetaBBO agent for automatically configuring diverse EAs across optimization problems. To this end, 
we first introduce a novel EA modularization system Modular-BBO that is capable of maintaining various sub-modules in EAs and spanning a massive algorithm structure space. We then formulate the universal AC over this algorithm space as an MTRL problem and hence propose meta-learning a Transformer based configuration policy to maximize the overall optimization performance across task samples. Extensive experiments demonstrate that a pre-trained ConfigX model could achieve superior AC performance to the state-of-the-art manual AC method SMAC3. Furthermore, we verify that ConfigX holds promising lifelong learning ability when being fine-tuned to adapt out-of-scope algorithm structures and optimization problems. We hope this work could serve as a pivotal step towards automatic and all-purpose AC base model. 

\section{Acknowledgments}
This work was supported in part by the National Natural Science Foundation of China (Grant No. 62276100), in part by the Guangdong Natural Science Funds for Distinguished Young Scholars (Grant No. 2022B1515020049), in part by the Guangdong Provincial Natural Science Foundation for Outstanding Youth Team Project (Grant No. 2024B1515040010), in part by the National Research Foundation, Singapore, under its AI Singapore Programme (AISG Award No. AISG3-RP-2022-031), and in part by the TCL Young Scholars Program.

\appendix
\section{List of Sub-modules}\label{appx:module}

We provide a complete list of the \textsc{Uncontrollable} and \textsc{Controllable} sub-modules included in Modular-BBO in Table~\ref{tab:uncontrol} and Table~\ref{tab:control} respectively, where the module types, concrete variants' names, functional descriptions, configuration space~(only for \textsc{Controllable}) and topology rules are presented. The configuration space description for each sub-module variant includes not only its legal value space but also its default value in the original paper. The topology rule of each sub-module variant, which contains a list of sub-module types, provides an information for algorithm structure generation procedure about which type of sub-modules can not be placed after this sub-module variant. Specially, we have only showcased a few \textsc{Multi\_Strategy} sub-module variants in the table due to the space limitation. For a \textsc{Multi\_Strategy} sub-module, it holds a more complicated configuration space than the others since it is an ensemble of other simple sub-module variants. When it shows up in a generated algorithm structure, our ConfigX has to select a sub-module variant in its ensemble first and configure the corresponding hyper-parameters at the same time. 

\begin{table*}[!ht]
\centering
\resizebox{\textwidth}{!}{%
\begin{tabular}{c|c|l|l}
\hline
\multirow{2}{*}{type}
& \multicolumn{3}{c}{Sub-module}
\\ \cline{2-4} 
& Name + Id
& Functional Description
& Topology Rule
\\ \hline
\multirow{11}{*}{\textsc{Initialization}}
& \begin{tabular}[c]{@{}c@{}}Uniform~\cite{initsurvey}\\0 - 000001 - 000000001\end{tabular}
& \begin{tabular}[l]{@{}l@{}}Uniformly sample solutions in the search range $x\sim\mathcal{U}(lb, ub)$\\where $ub$ and $lb$ are the upper and lower bounds of the search space.\end{tabular}
& \begin{tabular}[l]{@{}l@{}}Legal followers: DE-style \textsc{Mutation}, \textsc{PSO\_Update},\\ GA-style \textsc{Crossover}, \textsc{Multi\_Strategy}\end{tabular}
\\ \cline{2-4} 
& \begin{tabular}[c]{@{}c@{}}Sobol~\cite{sobol2008}\\0 - 000001 - 000000010\end{tabular}
& Sample population in Sobol’ sequences.
& \begin{tabular}[l]{@{}l@{}}Legal followers: DE-style \textsc{Mutation}, \textsc{PSO\_Update},\\ GA-style \textsc{Crossover}, \textsc{Multi\_Strategy}\end{tabular}
\\ \cline{2-4} 
& \begin{tabular}[c]{@{}c@{}}LHS~\cite{LHS}\\0 - 000001 - 000000011\end{tabular}
& Sample population in Latin hypercube sampling.
& \begin{tabular}[l]{@{}l@{}}Legal followers: DE-style \textsc{Mutation}, \textsc{PSO\_Update},\\ GA-style \textsc{Crossover}, \textsc{Multi\_Strategy}\end{tabular}
\\ \cline{2-4} 
& \begin{tabular}[c]{@{}c@{}}Halton~\cite{Halton}\\0 - 000001 - 000000100\end{tabular}
& Sample population in Halton sequence.
& \begin{tabular}[l]{@{}l@{}}Legal followers: DE-style \textsc{Mutation}, \textsc{PSO\_Update},\\ GA-style \textsc{Crossover}, \textsc{Multi\_Strategy}\end{tabular}
\\ \cline{2-4} 
& \begin{tabular}[c]{@{}c@{}}Normal~\cite{mahdavi2016center}\\0 - 000001 - 000000101\end{tabular}
& \begin{tabular}[l]{@{}l@{}}Sample solutions in Normal distribution $x\sim\mathcal{N}((ub+lb)/2, \frac{1}{6}(ub-lb))$\\ where $ub$ and $lb$ are the upper and lower bounds of the search space.\end{tabular}
& \begin{tabular}[l]{@{}l@{}}Legal followers: DE-style \textsc{Mutation}, \textsc{PSO\_Update},\\ GA-style \textsc{Crossover}, \textsc{Multi\_Strategy}\end{tabular}
\\ \hline
\multirow{3}{*}{\textsc{Niching}}
& \begin{tabular}[c]{@{}c@{}}Rand~\cite{dmspso}\\0 - 000010 - 000000001\end{tabular}
& \begin{tabular}[l]{@{}l@{}}Randomly partition the overall population into $N_{nich}\in[2,4]$ same size\\ sub-populations.\end{tabular}
& \begin{tabular}[l]{@{}l@{}}Legal followers: DE-style \textsc{Mutation}, \textsc{PSO\_Update},\\ GA-style \textsc{Crossover}, \textsc{Multi\_Strategy}\end{tabular}
\\ \cline{2-4} 
& \begin{tabular}[c]{@{}c@{}}Ranking~\cite{arruda2008pid}\\0 - 000010 - 000000010\end{tabular}
& \begin{tabular}[l]{@{}l@{}}Sort the population according to their fitness and partition them into \\ $N_{nich}\in[2,4]$ same size sub-populations.\end{tabular}
& \begin{tabular}[l]{@{}l@{}}Legal followers: DE-style \textsc{Mutation}, \textsc{PSO\_Update},\\ GA-style \textsc{Crossover}, \textsc{Multi\_Strategy}\end{tabular}
\\ \cline{2-4} 
& \begin{tabular}[c]{@{}c@{}}Distance~\cite{liu2020niching}\\0 - 000010 - 000000011\end{tabular}
& \begin{tabular}[l]{@{}l@{}}Randomly select a solution and assign its $NP//N_{nich} -1 $ nearest solutions\\ to a new sub-population, until all solutions are assigned.\end{tabular}
& \begin{tabular}[l]{@{}l@{}}Legal followers: DE-style \textsc{Mutation}, \textsc{PSO\_Update},\\ GA-style \textsc{Crossover}, \textsc{Multi\_Strategy}\end{tabular}
\\ \hline
\multirow{3}{*}{\textsc{Boundary\_Control}}
& \begin{tabular}[c]{@{}c@{}}Clip~\cite{kadavy2023impact}\\0 - 000011 - 000000001\end{tabular}
& \begin{tabular}[l]{@{}l@{}}Clip the solutions out of bounds at the bound $x_i =\text{clip}(x_i, lb, ub)$\end{tabular}
& \begin{tabular}[l]{@{}l@{}}Legal followers: \textsc{Selection}\end{tabular}
\\ \cline{2-4} 
& \begin{tabular}[c]{@{}c@{}}Rand~\cite{kadavy2023impact}\\0 - 000011 - 000000010\end{tabular}
& \begin{tabular}[l]{@{}l@{}}Randomly regenerate those out of bounds $x_{i,j}=\begin{cases}
    x_{i,j}, \text{ if } lb_j \leq x_{i,j} \leq ub_j,\\
    \mathcal{U}(lb_j,ub_j), \text{ otherwise}
\end{cases}$.\end{tabular}
& \begin{tabular}[l]{@{}l@{}}Legal followers: \textsc{Selection}\end{tabular}
\\ \cline{2-4} 
& \begin{tabular}[c]{@{}c@{}}Periodic~\cite{kadavy2023impact}\\0 - 000011 - 000000011\end{tabular}
& \begin{tabular}[l]{@{}l@{}}Consider the search range as a closed loop \\$x_{i,j}=\begin{cases}
    x_{i,j}, \text{ if } lb_j \leq x_{i,j} \leq ub_j,\\
    lb_j+((x_{i,j} - ub_j) \mod (ub_j - lb_j)), \text{ otherwise}
\end{cases}$.\end{tabular}
& \begin{tabular}[l]{@{}l@{}}Legal followers: \textsc{Selection}\end{tabular}
\\ \cline{2-4} 
& \begin{tabular}[c]{@{}c@{}}Reflect~\cite{kadavy2023impact}\\0 - 000011 - 000000100\end{tabular}
& \begin{tabular}[l]{@{}l@{}}Reflect the values that hit the bound  $x_{i,j}=\begin{cases}
    2ub_j - x_{i,j}, \text{ if } ub_j < x_{i,j},\\
    2lb_j - x_{i,j}, \text{ if } x_{i,j} < lb_j,\\
    x_{i,j}, \text{ otherwise}
\end{cases}$\end{tabular}
& \begin{tabular}[l]{@{}l@{}}Legal followers: \textsc{Selection}\end{tabular}
\\ \cline{2-4} 
& \begin{tabular}[c]{@{}c@{}}Halving~\cite{kadavy2023impact}\\0 - 000011 - 000000101\end{tabular}
& \begin{tabular}[l]{@{}l@{}}Halve the distance between the $x_i$ and the crossed bound\\  $x_{i,j}=\begin{cases}
    x_{i,j} + 0.5\cdot(x_{i,j} - ub_j), \text{ if } ub_j < x_{i,j},\\
    x_{i,j} + 0.5\cdot(x_{i,j} - lb_j), \text{ if } x_{i,j} < lb_j,\\
    x_{i,j}, \text{ otherwise}
\end{cases}$\end{tabular}
& \begin{tabular}[l]{@{}l@{}}Legal followers: \textsc{Selection}\end{tabular}
\\ \hline
\multirow{13}{*}{\textsc{Selection}}
& \begin{tabular}[c]{@{}c@{}}DE-like~\cite{DE}\\0 - 000100 - 000000001\end{tabular}
& \begin{tabular}[l]{@{}l@{}}Select the better one from the parent solution and its trail solution.\end{tabular}
& \begin{tabular}[l]{@{}l@{}}Legal followers: \textsc{Restart\_Strategy}, \\\textsc{Population\_Reduction}, \textsc{Completed}, \\\textsc{Information\_Sharing} (If \textsc{Niching} is used)\end{tabular}
\\ \cline{2-4} 
& \begin{tabular}[c]{@{}c@{}}Crowding~\cite{jDE21}\\0 - 000100 - 000000010\end{tabular}
& \begin{tabular}[l]{@{}l@{}}The trail solution complete against its closest solution\\ and the better one survives.\end{tabular}
& \begin{tabular}[l]{@{}l@{}}Legal followers: \textsc{Restart\_Strategy}, \\\textsc{Population\_Reduction}, \textsc{Completed}, \\\textsc{Information\_Sharing} (If \textsc{Niching} is used)\end{tabular}
\\ \cline{2-4} 
& \begin{tabular}[c]{@{}c@{}}PSO-like~\cite{PSO}\\0 - 000100 - 000000011\end{tabular}
& \begin{tabular}[l]{@{}l@{}}Replace the old population with the new solutions\\ without objective value comparisons.\end{tabular}
& \begin{tabular}[l]{@{}l@{}}Legal followers: \textsc{Restart\_Strategy}, \\\textsc{Population\_Reduction}, \textsc{Completed}, \\\textsc{Information\_Sharing} (If \textsc{Niching} is used)\end{tabular}
\\ \cline{2-4} 
& \begin{tabular}[c]{@{}c@{}}Ranking~\cite{baker2014adaptive}\\0 - 000100 - 000000100\end{tabular}
& \begin{tabular}[l]{@{}l@{}}Select solutions for the next generation according to the ranking based\\ probabilities, with the worst one ranking 1, the probability of the solution\\ rank $i$ is 
$p_i = \frac{1}{NP}(p^- + (p^+ - p^-)\frac{i-1}{NP - 1})$ where $NP$ is the population\\ size, $p^+$ is the probability of selecting the best solution and $p^-$ is the\\ probability of selecting the worst one.\end{tabular}
& \begin{tabular}[l]{@{}l@{}}Legal followers: \textsc{Restart\_Strategy}, \\\textsc{Population\_Reduction}, \textsc{Completed}, \\\textsc{Information\_Sharing} (If \textsc{Niching} is used)\end{tabular}
\\ \cline{2-4} 
& \begin{tabular}[c]{@{}c@{}}Tournament~\cite{goldberg1991comparative}\\0 - 000100 - 000000101\end{tabular}
& \begin{tabular}[l]{@{}l@{}}Randomly pair solutions and select the better\\ one in each pair for the next generation.\end{tabular}
& \begin{tabular}[l]{@{}l@{}}Legal followers: \textsc{Restart\_Strategy}, \\\textsc{Population\_Reduction}, \textsc{Completed}, \\\textsc{Information\_Sharing} (If \textsc{Niching} is used)\end{tabular}
\\ \cline{2-4} 
& \begin{tabular}[c]{@{}c@{}}Roulette~\cite{GA}\\0 - 000100 - 000000110\end{tabular}
& \begin{tabular}[l]{@{}l@{}}Select solutions according to the fitness based probabilities $p_i= \frac{f'_i}{\sum_{j=1}^{NP} f'_j}$ \\where $f'_j$ is the fitness of the $j$-th solution and $NP$ is population size.\end{tabular}
& \begin{tabular}[l]{@{}l@{}}Legal followers: \textsc{Restart\_Strategy}, \\\textsc{Population\_Reduction}, \textsc{Completed}, \\\textsc{Information\_Sharing} (If \textsc{Niching} is used)\end{tabular}
\\ \hline
\multirow{8}{*}{\textsc{Restart\_Strategy}}
& \begin{tabular}[c]{@{}c@{}}Stagnation~\cite{rJADE}\\0 - 000101 - 000000001\end{tabular}
& \begin{tabular}[l]{@{}l@{}}Reinitialize the population if the improvement of the best objective\\ value is equal to or less than a threshold $10^{-10}$ for 100 generations. \end{tabular}
& \begin{tabular}[l]{@{}l@{}}Legal followers: \textsc{Completed}\end{tabular}
\\ \cline{2-4}
& \begin{tabular}[c]{@{}c@{}}Obj\_Convergence~\cite{jDE21}\\0 - 000101 - 000000010\end{tabular}
& \begin{tabular}[l]{@{}l@{}}Reinitialize the population if the maximal difference of the objective\\ values of the top 20\% solutions is less than a threshold $10^{-16}$.\end{tabular}
& \begin{tabular}[l]{@{}l@{}}Legal followers: \textsc{Completed}\end{tabular}
\\ \cline{2-4}
& \begin{tabular}[c]{@{}c@{}}Solution\_Convergence~\cite{zhabitskaya2013asynchronous}\\0 - 000101 - 000000011\end{tabular}
& \begin{tabular}[l]{@{}l@{}}Reinitialize the population if the maximal difference of the solutions\\ on all dimensions are less than a threshold $10^{-16}$ search space diameter.  \end{tabular}
& \begin{tabular}[l]{@{}l@{}}Legal followers: \textsc{Completed}\end{tabular}
\\ \cline{2-4}
& \begin{tabular}[c]{@{}c@{}}Obj\&Solution\_Convergence~\cite{polakova2014controlled}\\0 - 000101 - 000000100\end{tabular}
& \begin{tabular}[l]{@{}l@{}}Reinitialize the population if the maximal difference of the objective\\ values is less than threshold $10^{-8}$ and the maximal distance among\\ solutions is less than 0.005 search space diameter.\end{tabular}
& \begin{tabular}[l]{@{}l@{}}Legal followers: \textsc{Completed}\end{tabular}
\\ \hline
\multirow{5}{*}{\textsc{Population\_Reduction}} 
& \begin{tabular}[c]{@{}c@{}}Linear~\cite{LSHADE}\\0 - 000110 - 000000001\end{tabular}
& \begin{tabular}[l]{@{}l@{}}Linearly reduce the population size from the initial size $NP_{max}$ to the\\ minimal population size $NP_{min}$. The size at generation $g+1$ is \\$NP_{g+1} = round((NP_{min}-NP_{max})\cdot\frac{g}{H}) + NP_{max}$ \\where $g$ is the generation number and $H$ is the optimization horizon.\end{tabular}
& \begin{tabular}[l]{@{}l@{}}Legal followers: Restart\_Strategy, \textsc{Completed}\end{tabular}
\\ \cline{2-4}
& \begin{tabular}[c]{@{}c@{}}Non-Linear~\cite{NL-SHADE-RSP}\\0 - 000110 - 000000010\end{tabular}
& \begin{tabular}[l]{@{}l@{}}Non-linearly determine the $g+1$ generation population size as\\$NP_{g+1} = round((NP_{min}-NP_{max})^{1-g/H} + NP_{max})$\\ where $NP_{min}$ and $NP_{max}$ are the minimal and maximal population\\ sizes, $g$ is the generation number and $H$ is the optimization horizon.\end{tabular}
& \begin{tabular}[l]{@{}l@{}}Legal followers: Restart\_Strategy, \textsc{Completed}\end{tabular}
\\ \hline
\multirow{1}{*}{\textsc{Completed}} 
& \begin{tabular}[c]{@{}c@{}}\textsc{Completed}\\0 - 000111 - 000000001\end{tabular}
& \begin{tabular}[l]{@{}l@{}}A token indicating the completion of algorithm structure generation \\which has no practical function.\end{tabular}
& \begin{tabular}[l]{@{}l@{}}--\end{tabular}
\\ \hline

\end{tabular}%
}
\caption{The list of the practical variants of \textsc{Uncontrollable} modules.}
\label{tab:uncontrol}
\end{table*}

\begin{table*}[!ht]
\centering
\resizebox{\textwidth}{!}{%
\begin{tabular}{c|c|l|l|l}
\hline
\multirow{2}{*}{type}
& \multicolumn{4}{c}{Sub-module}
\\ \cline{2-5} 
& Name + Id
& Functional Description
& Configuration Space
& Topology Rule
\\ \hline
\multirow{3}{*}{\textsc{Mutation}}
& \begin{tabular}[c]{@{}c@{}}DE/rand/1~\cite{DE}\\1 - 000001 - 000000001\end{tabular}
& \begin{tabular}[l]{@{}l@{}}Generate solution $x_i$'s trail solution $v_i = x_{r1} + F1\cdot(x_{r2} - x_{r3})$\\ where $x_{r\cdot}$ are randomly selected solutions.\end{tabular}
& \begin{tabular}[l]{@{}l@{}}$F1 \in [0,1]$, default to 0.5.\end{tabular}
& \begin{tabular}[l]{@{}l@{}}Legal followers: DE-style \textsc{Crossover}, \textsc{Multi\_Strategy}\end{tabular}
\\ \cline{2-5} 
& \begin{tabular}[c]{@{}c@{}}DE/rand/2~\cite{DE}\\1 - 000001 - 000000010\end{tabular}
& \begin{tabular}[l]{@{}l@{}}Generate solution $x_i$'s trail solution by $v_i = x_{r1} + F1\cdot(x_{r2} - x_{r3}) + F2\cdot(x_{r4} - x_{r5})$\\ where $x_{r\cdot}$ are randomly selected solutions.\end{tabular}
& \begin{tabular}[l]{@{}l@{}}$F1,F2 \in [0,1]$, default to 0.5.\\\end{tabular}
& \begin{tabular}[l]{@{}l@{}}Legal followers: DE-style \textsc{Crossover}, \textsc{Multi\_Strategy}\end{tabular}
\\ \cline{2-5} 
& \begin{tabular}[c]{@{}c@{}}DE/best/1~\cite{DE}\\1 - 000001 - 000000011\end{tabular}
& \begin{tabular}[l]{@{}l@{}}Generate solution $x_i$'s trail solution by $v_i = x_{best} + F1\cdot(x_{r1} - x_{r2})$\\ where $x_{r\cdot}$ are randomly selected solutions and $x_{best}$ is the best solution.\end{tabular}
& \begin{tabular}[l]{@{}l@{}}$F1 \in [0,1]$, default to 0.5.\end{tabular}
& \begin{tabular}[l]{@{}l@{}}Legal followers: DE-style \textsc{Crossover}, \textsc{Multi\_Strategy}\end{tabular}
\\ \cline{2-5} 
& \begin{tabular}[c]{@{}c@{}}DE/best/2~\cite{DE}\\1 - 000001 - 000000100\end{tabular}
& \begin{tabular}[l]{@{}l@{}}Generate solution $x_i$'s trail solution by $v_i = x_{best} + F1\cdot(x_{r1} - x_{r2}) + F2\cdot(x_{r3} - x_{r4})$\\ where $x_{r\cdot}$ are randomly selected solutions and $x_{best}$ is the best solution.\end{tabular}
& \begin{tabular}[l]{@{}l@{}}$F1,F2 \in [0,1]$, default to 0.5.\\\end{tabular}
& \begin{tabular}[l]{@{}l@{}}Legal followers: DE-style \textsc{Crossover}, \textsc{Multi\_Strategy}\end{tabular}
\\ \cline{2-5} 
& \begin{tabular}[c]{@{}c@{}}DE/current-to-best/1~\cite{DE}\\1 - 000001 - 000000101\end{tabular}
& \begin{tabular}[l]{@{}l@{}}Generate solution $x_i$'s trail solution by $v_i = x_{i} + F1\cdot(x_{best} - x_{i}) + F2\cdot(x_{r1} - x_{r2})$\\ where $x_{r\cdot}$ are randomly selected solutions and $x_{best}$ is the best solution.\end{tabular}
& \begin{tabular}[l]{@{}l@{}}$F1,F2 \in [0,1]$, default to 0.5.\\\end{tabular}
& \begin{tabular}[l]{@{}l@{}}Legal followers: DE-style \textsc{Crossover}, \textsc{Multi\_Strategy}\end{tabular}
\\ \cline{2-5} 
& \begin{tabular}[c]{@{}c@{}}DE/current-to-rand/1~\cite{DE}\\1 - 000001 - 000000110\end{tabular}
& \begin{tabular}[l]{@{}l@{}}Generate solution $x_i$'s trail solution by $v_i = x_{i} + F1\cdot(x_{r1} - x_{i}) + F2\cdot(x_{r2} - x_{r3})$\\ where $x_{r\cdot}$ are randomly selected solutions.\end{tabular}
& \begin{tabular}[l]{@{}l@{}}$F1,F2 \in [0,1]$, default to 0.5.\\\end{tabular}
& \begin{tabular}[l]{@{}l@{}}Legal followers: DE-style \textsc{Crossover}, \textsc{Multi\_Strategy}\end{tabular}
\\ \cline{2-5} 
& \begin{tabular}[c]{@{}c@{}}DE/rand-to-best/1~\cite{DE}\\1 - 000001 - 000000111\end{tabular}
& \begin{tabular}[l]{@{}l@{}}Generate solution $x_i$'s trail solution by $v_i = x_{r1} + F1\cdot(x_{best} - x_{r2})$\\ where $x_{r\cdot}$ are randomly selected solutions and $x_{best}$ is the best solution.\end{tabular}
& \begin{tabular}[l]{@{}l@{}}$F1 \in [0,1]$, default to 0.5.\\\end{tabular}
& \begin{tabular}[l]{@{}l@{}}Legal followers: DE-style \textsc{Crossover}, \textsc{Multi\_Strategy}\end{tabular}
\\ \cline{2-5} 
& \begin{tabular}[c]{@{}c@{}}DE/current-to-pbest/1~\cite{JADE}\\1 - 000001 - 000001000\end{tabular}
& \begin{tabular}[l]{@{}l@{}}Generate solution $x_i$'s trail solution by $v_i = x_{i} + F1\cdot(x_{pbest} - x_{i}) + F2\cdot(x_{r1} - x_{r2})$\\ where $x_{r\cdot}$ are randomly selected solutions and $x_{pbest}$ is a randomly selected from the\\ top $p$ best solutions.\end{tabular}
& \begin{tabular}[l]{@{}l@{}}$F1,F2 \in [0,1]$, default to 0.5;\\$p\in[0, 1]$, default to 0.05.\\\end{tabular}
& \begin{tabular}[l]{@{}l@{}}Legal followers: DE-style \textsc{Crossover}, \textsc{Multi\_Strategy}\end{tabular}
\\ \cline{2-5} 
& \begin{tabular}[c]{@{}c@{}}DE/current-to-pbest/1+archive~\cite{JADE}\\1 - 000001 - 000001001\end{tabular}
& \begin{tabular}[l]{@{}l@{}}Generate solution $x_i$'s trail solution by $v_i = x_{i} + F1\cdot(x_{pbest} - x_{i}) + F2\cdot(x_{r1} - x_{r2})$\\ where $x_{r1}$ is a randomly selected solutions, $x_{r2}$ is randomly selected from the union of\\ the population and the archive which contains inferior solutions, $x_{pbest}$ is a randomly\\ selected solution from the top $p$ best solutions.\end{tabular}
& \begin{tabular}[l]{@{}l@{}}$F1,F2 \in [0,1]$, default to 0.5;\\$p\in[0, 1]$, default to 0.05.\\\end{tabular}
& \begin{tabular}[l]{@{}l@{}}Legal followers: DE-style \textsc{Crossover}, \textsc{Multi\_Strategy}\end{tabular}
\\ \cline{2-5} 
& \begin{tabular}[c]{@{}c@{}}DE/weighted-rand-to-pbest/1~\cite{MadDE}\\1 - 000001 - 000001010\end{tabular}
& \begin{tabular}[l]{@{}l@{}}Generate solution $x_i$'s trail solution by $v_i = F1\cdot x_{r1} + F1\cdot F2\cdot(x_{pbest} - x_{r2})$\\ where $x_{r\cdot}$ are randomly selected solutions and $x_{best}$ is the best solution.\end{tabular}
& \begin{tabular}[l]{@{}l@{}}$F1,F2 \in [0,1]$, default to 0.5;\\$p\in[0, 1]$, default to 0.05.\\\end{tabular}
& \begin{tabular}[l]{@{}l@{}}Legal followers: DE-style \textsc{Crossover}, \textsc{Multi\_Strategy}\end{tabular}
\\ \cline{2-5} 
& \begin{tabular}[c]{@{}c@{}}DE/current-to-rand/1+archive~\cite{MadDE}\\1 - 000001 - 000001011\end{tabular}
& \begin{tabular}[l]{@{}l@{}}Generate solution $x_i$'s trail solution by $v_i = x_{i} + F1\cdot(x_{r1} - x_{i}) + F2\cdot(x_{r2} - x_{r3})$\\ where $x_{r1},x_{r2}$ are randomly selected solutions, $x_{r3}$ is randomly selected from the union\\ of the population and the archive which contains inferior solutions.\end{tabular}
& \begin{tabular}[l]{@{}l@{}}$F1,F2 \in [0,1]$, default to 0.5.\\\end{tabular}
& \begin{tabular}[l]{@{}l@{}}Legal followers: DE-style \textsc{Crossover}, \textsc{Multi\_Strategy}\end{tabular}
\\ \cline{2-5} 
& \begin{tabular}[c]{@{}c@{}}Gaussian\_mutation~\cite{GA}\\1 - 000001 - 000001100\end{tabular}
& \begin{tabular}[l]{@{}l@{}}Generate a mutated solution of $x_i$ by adding a Gaussian noise on each dimension\\ $v_{i} =  \mathcal{N}(x_{i}, \sigma\cdot(ub-lb))$ where $ub$ and $lb$ are the upper and lower bounds of the\\ search space.\end{tabular}
& \begin{tabular}[l]{@{}l@{}}$\sigma\in [0, 1]$, default to 0.1\end{tabular}
& \begin{tabular}[l]{@{}l@{}}Legal followers: \textsc{Boundary\_control}, \textsc{Multi\_Strategy}\end{tabular}
\\ \cline{2-5} 
& \begin{tabular}[c]{@{}c@{}}Polynomial\_mutation~\cite{dobnikar1999niched}\\1 - 000001 - 000001101\end{tabular}
& \begin{tabular}[l]{@{}l@{}}Generate a mutated solution of $x_i$ as $v_i = \begin{cases}
    x_i + ((2u)^{\frac{1}{1+\eta_m}} - 1)(x_i-lb), \text{if }u \leq 0.5;\\
    x_i + (1 - (2-2u)^{\frac{1}{1+\eta_m}})(ub - x_i),  \text{if }u > 0.5.
\end{cases}$ \\where $u\in[0,1]$ is a random number, $ub$ and $lb$ are the upper and lower bounds of the\\ search space.\end{tabular}
& \begin{tabular}[l]{@{}l@{}}$\eta_m \in [20,100]$, default to 20\end{tabular}
& \begin{tabular}[l]{@{}l@{}}Legal followers: \textsc{Boundary\_control}, \textsc{Multi\_Strategy}\end{tabular}
\\ \hline
\multirow{3}{*}{\textsc{Crossover}}
& \begin{tabular}[c]{@{}c@{}}Binomial~\cite{DE}\\1 - 000010 - 000000001\end{tabular}
& \begin{tabular}[l]{@{}l@{}}Randomly exchange values between parent solution $x_i$ and the trail solution $v_i$ to get a new solution:\\ $u_{i,j} = \begin{cases}
    v_{i,j}, \text{ if }rand_j < Cr \text{ or } j = jrand\\
    x_{i,j}, \text{ otherwise}
\end{cases}, j = 1,\cdots,D$ where $rand_j \in [0,1]$ is a\\ random number, $jrand\in[1,D]$ is a randomly selected index before crossover and $D$ is the\\ solution dimension.\end{tabular}
& \begin{tabular}[l]{@{}l@{}}$Cr\in[0,1]$, default to 0.9.\end{tabular}
& \begin{tabular}[l]{@{}l@{}}Legal followers: \textsc{Boundary\_control}, \textsc{Multi\_Strategy}\end{tabular}
\\ \cline{2-5} 
& \begin{tabular}[c]{@{}c@{}}Exponential~\cite{DE}\\1 - 000010 - 000000010\end{tabular}
& \begin{tabular}[l]{@{}l@{}}Exchange a random solution segment between $x_i$ and $v_i$ to get a new solution:\\$u_{i,j} = \begin{cases}
    v_{i,j}, \text{ if } rand_{k:j} < Cr \text{ and } k \leq j \leq L+k\\
    x_{i,j}, \text{ otherwise}
\end{cases}, j=1,\cdots,D$ where $k\in[1,D]$ is a randomly\\ selected start point for exchanging, $L\in[1,D-k]$ is a randomly determined exchange length, \\$rand_{k:j} \in [0,1]^{j-k}$ is the random numbers from index $k$ to $j$ and $D$ is the solution dimension. \end{tabular}
& \begin{tabular}[l]{@{}l@{}}$Cr\in[0,1]$, default to 0.9.\end{tabular}
& \begin{tabular}[l]{@{}l@{}}Legal followers: \textsc{Boundary\_control}, \textsc{Multi\_Strategy}\end{tabular}
\\ \cline{2-5} 
& \begin{tabular}[c]{@{}c@{}}qbest\_Binomial~\cite{islam2011adaptive}\\1 - 000010 - 000000011\end{tabular}
& \begin{tabular}[l]{@{}l@{}}Randomly exchange values between a solution $x'_i$ selected from the top $p$ population and the trail\\ solution $v_i$ to get a new solution:\\ $u_{i,j} = \begin{cases}
    v_{i,j}, \text{ if }rand_j < Cr \text{ or } j = jrand\\
    x'_{i,j}, \text{ otherwise}
\end{cases}, j = 1,\cdots,D$ where $rand_j \in [0,1]$ is a\\ random number, $jrand\in[1,D]$ is a randomly selected index before crossover and $D$ is the\\ solution dimension.\end{tabular}
& \begin{tabular}[l]{@{}l@{}}$Cr\in[0,1]$, default to 0.9;\\$p\in[0,1]$, default to 0.5\end{tabular}
& \begin{tabular}[l]{@{}l@{}}Legal followers: \textsc{Boundary\_control}, \textsc{Multi\_Strategy}\end{tabular}
\\ \cline{2-5} 
& \begin{tabular}[c]{@{}c@{}}qbest\_Binomial+archive~\cite{MadDE}\\1 - 000010 - 000000100\end{tabular}
& \begin{tabular}[l]{@{}l@{}}Randomly exchange values between a solution $x'_i$ selected from the top $p$ population-archive union\\ and the trail solution $v_i$ to get a new solution:\\ $u_{i,j} = \begin{cases}
    v_{i,j}, \text{ if }rand_j < Cr \text{ or } j = jrand\\
    x'_{i,j}, \text{ otherwise}
\end{cases}, j = 1,\cdots,D$ where $rand_j \in [0,1]$ is a\\ random number, $jrand\in[1,D]$ is a randomly selected index before crossover and $D$ is the\\ solution dimension.\end{tabular}
& \begin{tabular}[l]{@{}l@{}}$Cr\in[0,1]$, default to 0.9;\\$p\in[0,1]$, default to 0.18\end{tabular}
& \begin{tabular}[l]{@{}l@{}}Legal followers: \textsc{Boundary\_control}, \textsc{Multi\_Strategy}\end{tabular}
\\ \cline{2-5} 
& \begin{tabular}[c]{@{}c@{}}SBX~\cite{sbx}\\1 - 000010 - 000000101\end{tabular}
& \begin{tabular}[l]{@{}l@{}}Generate child solution(s) $v_i$ by $v_i = 0.5\cdot [(1 \mp \beta) x_{p1} + (1 \pm \beta) x_{p2}]$ \\where $\beta = \begin{cases}
    (2u)^{\frac{1}{1+\eta_c}} - 1, \text{if }u \leq 0.5;\\
    (\frac{1}{2-2u})^{\frac{1}{1+\eta_c}},  \text{if }u > 0.5.
\end{cases}$, $u\in[0,1]$ is a random number, $x_{p1}$ and $x_{p2}$ are two\\ randomly selected parents.
\end{tabular}
& \begin{tabular}[l]{@{}l@{}}$\eta_c \in [20,100]$, default to 20\end{tabular}
& \begin{tabular}[l]{@{}l@{}}Legal followers: GA-style \textsc{Mutation}, \textsc{Multi\_Strategy}\end{tabular}
\\ \cline{2-5} 
& \begin{tabular}[c]{@{}c@{}}Arithmetic~\cite{michalewicz2013genetic}\\1 - 000010 - 000000110\end{tabular}
& \begin{tabular}[l]{@{}l@{}}Generate child solution $v_i$ by $v_i = (1 - \alpha)\cdot x_{p1} + \alpha\cdot x_{p2}$ where $x_{p1}$ and $x_{p2}$ are two\\ randomly selected parents.\end{tabular}
& \begin{tabular}[l]{@{}l@{}}$\alpha\in[0,1]$, default to 0.5.\end{tabular}
& \begin{tabular}[l]{@{}l@{}}Legal followers: GA-style \textsc{Mutation}, \textsc{Multi\_Strategy}\end{tabular}
\\ \hline
\multirow{3}{*}{\textsc{PSO\_Update}}
& \begin{tabular}[c]{@{}c@{}}Vanilla\_PSO~\cite{PSO}\\1 - 000011 - 000000001\end{tabular}
& \begin{tabular}[l]{@{}l@{}}Update solution $x^t_i$ at generation $t$ using $x^{t+1}_i = x^t_i + vel^t_i$ where velocity vector \\$vel^t_i = w\cdot vel^{t-1}_i + c1\cdot rand_1 \cdot (pbest^t_i - x^t_i) + c2\cdot rand_2 \cdot (gbest^t - x^t_i)$, \\$rand_\cdot \in (0,1]$ are random values, $pbest^t_i$ is the best solution $x_i$ ever achieved, $gbest^t$ is the \\global best solution.\end{tabular}
& \begin{tabular}[l]{@{}l@{}}$w\in [0.4,0.9]$, default to 0.7;\\$c1,c2 \in [0,2]$, default to 1.49445.\end{tabular}
& \begin{tabular}[l]{@{}l@{}}Legal followers: \textsc{Boundary\_control}, \textsc{Multi\_Strategy}\end{tabular}
\\ \cline{2-5} 
& \begin{tabular}[c]{@{}c@{}}FDR\_PSO~\cite{FDRPSO}\\1 - 000011 - 000000010\end{tabular}
& \begin{tabular}[l]{@{}l@{}}Update solution $x^t_i$ at generation $t$ using $x^{t+1}_i = x^t_i + vel^t_i$ where velocity vector \\$vel^t_i = w\cdot vel^{t-1}_i + c1\cdot rand_1 \cdot (pbest^t_i - x_i) + c2\cdot rand_2 \cdot (gbest^t - x_i) + c3\cdot rand_3 \cdot (nbest^t_i - x_i)$, \\$rand_\cdot \in (0,1]$ are random values, $pbest^t_i$ is the best solution $x_i$ ever achieved, $gbest^t$ is the global best\\ solution and $nbest^t_i$ is the solution that maximizes the Fitness-Distance-Ratio $nbest^t_{i,j} = x^t_{p_j,j}$ which\\ $p_j = \mathop{\arg\max}\limits_{p\in[1,NP]}\frac{f^t_i - f^t_{p_j}}{|x^t_{p_j,j} - x^t_{i,j}|}, j=1,\cdots,D$, $f$ denotes the objective values and $D$ is solution dimension. \end{tabular}
& \begin{tabular}[l]{@{}l@{}}$w\in [0.4,0.9]$, default to 0.729;\\$c1,c2 \in [0,2]$, default to 1;\\$c3\in[0,2]$, default to 2.\end{tabular}
& \begin{tabular}[l]{@{}l@{}}Legal followers: \textsc{Boundary\_control}, \textsc{Multi\_Strategy}\end{tabular}
\\ \cline{2-5} 
& \begin{tabular}[c]{@{}c@{}}CLPSO~\cite{CLPSO}\\1 - 000011 - 000000011\end{tabular}
& \begin{tabular}[l]{@{}l@{}}Update solution $x^t_i$ at generation $t$ using $x^{t+1}_i = x^t_i + vel^t_i$ where velocity vector \\$vel^t_i = w\cdot vel^{t-1}_i + c1\cdot rand_1 \cdot (pbest^t_{f_i} - x^t_i) + c2\cdot rand_2 \cdot (gbest^t - x^t_i)$, where $rand_\cdot \in (0,1]$ are \\random values, $gbest^t$ is the global best solution, $pbest^t_{f_i,j} = \begin{cases}
    pbest^t_{i,j}, \text{ if } rand_j > Pc_i;\\
    pbest^t_{r,j}, \text{ otherwise.}
\end{cases}, j=1,\cdots,D$ \\is the ever achieved best solution of $x_i$ or $x_r$ which is randomly selected with fitness based tournament.\end{tabular}
& \begin{tabular}[l]{@{}l@{}}$w\in [0.4,0.9]$, default to 0.7;\\$c1,c2 \in [0,2]$, default to 1.49445.\end{tabular}
& \begin{tabular}[l]{@{}l@{}}Legal followers: \textsc{Boundary\_control}, \textsc{Multi\_Strategy}\end{tabular}
\\ \hline
\multirow{3}{*}{\textsc{Multi\_Strategy}}
& \begin{tabular}[c]{@{}c@{}}Multi\_Niching\_2\\1 - 000100 - 000000001\end{tabular}
& \begin{tabular}[l]{@{}l@{}}It contains \textsc{Rand}, \textsc{Ranking} and \textsc{Distance} three niching methods with the same sub-population size\\ $N_{nich}=2$, its action is to select one of the three methods to conduct niching.\end{tabular}
& \begin{tabular}[l]{@{}l@{}}$op\in\{\textsc{Rand}, \textsc{Ranking}, \textsc{Distance}\}$, \\default to \textsc{Rand}.\end{tabular}
& \begin{tabular}[l]{@{}l@{}}Legal followers: DE-style \textsc{Mutation}, \textsc{PSO\_Update},\\ GA-style \textsc{Crossover}, \textsc{Multi\_Strategy}\end{tabular}
\\ \cline{2-5} 
& \begin{tabular}[c]{@{}c@{}}Multi\_Niching\_3\\1 - 000100 - 000000010\end{tabular}
& \begin{tabular}[l]{@{}l@{}}It contains \textsc{Rand}, \textsc{Ranking} and \textsc{Distance} three niching methods with the same sub-population size\\ $N_{nich}=3$, its action is to select one of the three methods to conduct niching.\end{tabular}
& \begin{tabular}[l]{@{}l@{}}$op\in\{\textsc{Rand}, \textsc{Ranking}, \textsc{Distance}\}$, \\default to \textsc{Rand}.\end{tabular}
& \begin{tabular}[l]{@{}l@{}}Legal followers: DE-style \textsc{Mutation}, \textsc{PSO\_Update},\\ GA-style \textsc{Crossover}, \textsc{Multi\_Strategy}\end{tabular}
\\ \cline{2-5} 
& \begin{tabular}[c]{@{}c@{}}Multi\_Niching\_4\\1 - 000100 - 000000011\end{tabular}
& \begin{tabular}[l]{@{}l@{}}It contains \textsc{Rand}, \textsc{Ranking} and \textsc{Distance} three niching methods with the same sub-population size\\ $N_{nich}=4$, its action is to select one of the three methods to conduct niching.\end{tabular}
& \begin{tabular}[l]{@{}l@{}}$op\in\{\textsc{Rand}, \textsc{Ranking}, \textsc{Distance}\}$, \\default to \textsc{Rand}.\end{tabular}
& \begin{tabular}[l]{@{}l@{}}Legal followers: DE-style \textsc{Mutation}, \textsc{PSO\_Update},\\ GA-style \textsc{Crossover}, \textsc{Multi\_Strategy}\end{tabular}
\\ \cline{2-5} 
& \begin{tabular}[c]{@{}c@{}}Multi\_BC\\1 - 000100 - 000000100\end{tabular}
& \begin{tabular}[l]{@{}l@{}}It contains the five Boundary\_Control methods \textsc{Clip}, \textsc{Rand}, \textsc{Periodic}, \textsc{Reflect} and \textsc{Halving}, its\\ action is to select one of the five methods.\end{tabular}
& \begin{tabular}[l]{@{}l@{}}$op\in$\{\textsc{Clip}, \textsc{Rand}, \textsc{Periodic}, \textsc{Reflect}, \\\textsc{Halving}\}, default to \textsc{Clip}.\end{tabular}
& \begin{tabular}[l]{@{}l@{}}Legal followers: \textsc{Selection}\end{tabular}
\\ \cline{2-5} 
& \begin{tabular}[c]{@{}c@{}}Multi\_Mutation\_1~\cite{MadDE}\\1 - 000100 - 000000101\end{tabular}
& \begin{tabular}[l]{@{}l@{}}Contains DE/current-to-pbest/1+archive, DE/current-to-rand/1+archive and DE/weighted-rand-to-best/1\\ three DE mutation sub-modules, its first configuration is to select one of the three mutations and the rest\\ configurations are to configured the selected operator. \end{tabular}
& \begin{tabular}[l]{@{}l@{}}$op\in$\{\textsc{DE/current-to-pbest/1+archive},\\ \textsc{DE/current-to-rand/1+archive},\\ \textsc{DE/weighted-rand-to-best/1}\}, \\random selection in default;\\$F1,F2 \in [0,1]$, default to 0.5;\\$p\in[0, 1]$, default to 0.18.\end{tabular}
& \begin{tabular}[l]{@{}l@{}}Legal followers: DE-style \textsc{Crossover}\end{tabular}
\\ \cline{2-5} 
& \begin{tabular}[c]{@{}c@{}}Multi\_Mutation\_2~\cite{CoDE}\\1 - 000100 - 000000110\end{tabular}
& \begin{tabular}[l]{@{}l@{}}Contains DE/rand/1, DE/rand/2 and DE/current-to-rand/1 three DE mutation sub-modules.\end{tabular}
& \begin{tabular}[l]{@{}l@{}}$op\in$\{\textsc{DE/rand/1}, \textsc{DE/rand/2},\\ \textsc{DE/current-to-rand/1}\}, \\random selection in default;\\$F1,F2 \in [0,1]$, default to 0.5;\end{tabular}
& \begin{tabular}[l]{@{}l@{}}Legal followers: DE-style \textsc{Crossover}\end{tabular}
\\ \cline{2-5} 
& \begin{tabular}[c]{@{}c@{}}Multi\_Mutation\_3~\cite{EPSDE}\\1 - 000100 - 000000111\end{tabular}
& \begin{tabular}[l]{@{}l@{}}Contains DE/rand/1, DE/best/2 and DE/current-to-rand/1 three DE mutation sub-modules.\end{tabular}
& \begin{tabular}[l]{@{}l@{}}$op\in$\{\textsc{DE/rand/1}, \textsc{DE/best/2},\\ \textsc{DE/current-to-rand/1}\}, \\random selection in default;\\$F1,F2 \in [0,1]$, default to 0.5;\end{tabular}
& \begin{tabular}[l]{@{}l@{}}Legal followers: DE-style \textsc{Crossover}\end{tabular}
\\ \cline{2-5} 
& \begin{tabular}[c]{@{}c@{}}Multi\_Crossover\_1~\cite{MadDE}\\1 - 000100 - 000001000\end{tabular}
& \begin{tabular}[l]{@{}l@{}}Contains Binomial and qbest\_Binomial+archive two DE crossover sub-modules.\end{tabular}
& \begin{tabular}[l]{@{}l@{}}$op\in$\{\textsc{Binomial},\\ \textsc{qbest\_Binomial+archive}\}, \\random selection in default;\\$Cr \in [0,1]$, default to 0.9;\end{tabular}
& \begin{tabular}[l]{@{}l@{}}Legal followers: \textsc{Boundary\_Control}\end{tabular}
\\ \cline{2-5} 
& \begin{tabular}[c]{@{}c@{}}Multi\_Crossover\_2~\cite{peng2021multi}\\1 - 000100 - 000001001\end{tabular}
& \begin{tabular}[l]{@{}l@{}}Contains Binomial and Exponential two DE crossover sub-modules.\end{tabular}
& \begin{tabular}[l]{@{}l@{}}$op\in$\{\textsc{Binomial}, \textsc{Exponential}\}, \\random selection in default;\\$Cr \in [0,1]$, default to 0.9;\end{tabular}
& \begin{tabular}[l]{@{}l@{}}Legal followers: \textsc{Boundary\_Control}\end{tabular}
\\ \cline{2-5} 
& \begin{tabular}[c]{@{}c@{}}Multi\_PSO\_1~\cite{EPSO}\\1 - 000100 - 000001010\end{tabular}
& \begin{tabular}[l]{@{}l@{}}Contains FDR\_PSO and CLPSO two PSO update sub-modules.\end{tabular}
& \begin{tabular}[l]{@{}l@{}}$op\in$\{\textsc{FDR\_PSO}, \textsc{CLPSO},\}, \\random selection in default;\\$w\in [0.4,0.9]$, default to 0.729;\\$c1,c2 \in [0,2]$, default to 1;\\$c3\in[0,2]$, default to 2.\end{tabular}
& \begin{tabular}[l]{@{}l@{}}Legal followers: \textsc{Boundary\_Control}\end{tabular}
\\ \cline{2-5} 
& \begin{tabular}[c]{@{}c@{}}50 more Multi-Strategies about Mutations, Crossovers and \\PSO\_Updates are omitted here since they are too many\\ to presenting them one by one.\\1 - 000100 - 000001011\\$\sim$1 - 000100 - 000111101\end{tabular}
& \begin{tabular}[l]{@{}l@{}}$\cdots$\end{tabular}
& \begin{tabular}[l]{@{}l@{}}$\cdots$\end{tabular}
& \begin{tabular}[l]{@{}l@{}}$\cdots$\end{tabular}

\\ \hline
\multirow{1}{*}{\textsc{Information\_Sharing}}
& \begin{tabular}[c]{@{}c@{}}Sharing\\1 - 000101 - 000000001\end{tabular}
& \begin{tabular}[l]{@{}l@{}}Receive the best solution from the target sub-population and replace the worst solution in \\current sub-population.\end{tabular}
& \begin{tabular}[l]{@{}l@{}}$target\in[1,N_{nich}]$, random selection in default\end{tabular}
& \begin{tabular}[l]{@{}l@{}}Legal followers: \textsc{Population\_Reduction}, \textsc{Completed}\end{tabular}
\\ \hline
\end{tabular}%
}
\caption{The list of the practical variants of \textsc{Controllable} modules.}
\label{tab:control}
\end{table*}

\section{Algorithm Structure Generation}\label{appx:alg_gen}

The actual algorithm structure generation procedure in Modular-BBO follows the Algorithm~1 in the main body. It always starts with randomly selecting an \textsc{Initialization} sub-module variant. Given the selected initialization variant, the next sub-module which follows is decided by checking the \textbf{COMPLETED} and \textbf{VIOLATED} conditions. In such case, for the currently selected one, we first check the topology rule list of the it. If the randomly selected follower is not in this list, \textbf{VIOLATED} is set to \emph{true}, else \emph{false}. When an legal follower is selected, we check the \textbf{COMPLETED} condition by simply checking if it is a \textsc{Completed} sub-module, which is specially designed by us to denote the end of an algorithm structure~(see the last row in Table~\ref{tab:uncontrol}). Specially, to select a legal subsequent sub-module variant for \textsc{Niching} sub-module, it would introduce several branches to represent multi-sub-population algorithm structure. Hence, the following generation of \textsc{Niching} sub-module is separated into branches. For each branch (sub-population), we follow the same selection procedure until a \textsc{Completed} sub-module is selected for the corresponding branch. To summarize, bounded by these two conditions, the algorithm structure generation procedure ensures that: (a) all possible algorithm structures follow the common sense in regular EAs. (b) the sub-modules of different EA types will not appear together in a legal algorithm structure.

\section{State Details}\label{appx:state}
In Section 3.3 of the paper, we encode a information pair for the $j$-th sub-module in algorithm $\mathcal{A}_i$ as $s_{i,j}:\{s_{i,j}^\text{id}, s_i^\text{opt}\}$. Here $s_{i,j}^\text{id}\in\{0,1\}^{16}$ is the 16 bit binary module id presented in Table~\ref{tab:uncontrol} and Table~\ref{tab:control}. The algorithm performance information $s_i^\text{opt}\in\mathbb{R}^9$ contains 9 optimization features which are summarized below:
\begin{enumerate}
    \item The first feature is the minimum objective value in the current (sub-)population indicating the achieved best performance of the current (sub-)population:
    \begin{equation}
        s_{i,1}^\text{opt}=\min \{\frac{f_i}{f^{0,*}-f^*}\}_{i\in [1,NP_{local}]}
    \end{equation}
    It is normalized by the difference between the best objective value at initial optimization $f^{0,*}$ step and the global optimal objective value of the optimization problem $f^*$, so that the scales of the features from different tasks are in the same level. which hence stabilizes the training. $NP_{local}$ is the (sub-)population size.
    \item The second one is the averaged normalized objective values in the current (sub-)population, indicating the average performance of the (sub-)population:
    \begin{equation}
        s_{i,2}^\text{opt}=\text{mean}\{\frac{f_i}{f^{0,*}-f^*}\}_{i\in [1,NP_{local}]}
    \end{equation}
    \item The variance of the normalized objective values in the current (sub-)population, indicating the variance and convergence of the (sub-)population:
    \begin{equation}
        s_{i,3}^\text{opt}=\text{std} \{\frac{f_i}{f^{0,*}-f^*}\}_{i\in [1,NP_{local}]}
    \end{equation}
    \item The next feature is the maximal distance between the solutions in (sub-)population, normalized by the diameter of the search space, measuring the convergence:
    \begin{equation}
        s_{i,4}^\text{opt}=\mathop{\max}\limits_{i,j \in[1,NP_{local}]} \frac{||x_i - x_j||_2}{||ub-lb||_2}
    \end{equation}
    where $ub$ and $lb$ are the upper and lower bounds of the search space.
    \item The dispersion difference~\cite{lunacek2006dispersion} feature is calculated as the difference of the maximal distance between the top 10\% solutions and the maximal distance between all solutions in (sub-)population:
    \begin{equation}
    \begin{split}
        s_{i,5}^\text{opt}=\mathop{\max}\limits_{i,j \in[1,10\% NP_{local}]} \frac{||x_i - x_j||_2}{||ub-lb||_2} \\- \mathop{\max}\limits_{i,j \in[1,NP_{local}]} \frac{||x_i - x_j||_2}{||ub-lb||_2}
    \end{split}
    \end{equation}
    It measures the funnelity of the problem landscape: a single funnel problem has a smaller dispersion difference while the multi-funnel landscape has larger value.
    \item The fitness distance correlation (FDC)~\cite{fdc} describes the complexity of the problem by evaluating the relationship between fitness value and the distance of the solution from the optimum. 
    \begin{equation}
        s_{i,6}^\text{opt}=\frac{\frac{1}{NP_{local}}\sum_{i=1}^{NP_{local}}{(f_i - \Bar{f})(d_i^* - \Bar{d}^*)}}{\text{var}(\{d_i^*\}_{i\in [1,NP_{local}]})\cdot\text{var}(\{f_i\}_{i\in [1,NP_{local}]})}
    \end{equation}
    where the $\Bar{f}$ is the averaged objective value in (sub-)population, $d_i^* = ||x_i-x^*||_2$ is the distance between $x_i$ and the best solution $x^*$, $\Bar{d}^* = \text{mean} \{d_i^*\}_{i\in [1,NP_{local}]}$ is the averaged distance,\\ var($\cdot$) is the variance.
    \item The found global best objective among all (sub-)populations, indicating the achieved best performance of the overall optimization:
    \begin{equation}
        s_{i,7}^\text{opt}=\min \{\frac{f_i}{f^{0,*}-f^*}\}_{i\in [1,NP]}
    \end{equation}
    \item This feature is the FDC feature for the overall population:
    \begin{equation}
        s_{i,8}^\text{opt}=\frac{\frac{1}{NP}\sum_{i=1}^{NP}{(f_i - \Bar{f})(d_i^* - \Bar{d}^*)}}{\text{var}(\{d_i^*\}_{i\in [1,NP]})\cdot\text{var}(\{f_i\}_{i\in [1,NP]})}
    \end{equation}
    \item The last feature is the remaining optimization budget, indicating the optimization progress:
    \begin{equation}
        s_{i,9}^\text{opt}=\frac{H-t}{H}
    \end{equation}
    where $H$ is the optimization horizon and $t$ is the current optimization step.
\end{enumerate}
Feature 1$\sim$6 measures the local optimization status in the sub-population the sub-modules belonging to. If there is no Niching and only one population, these features measures the status of the global population. Features 7$\sim$9 describes the global optimization across sub-populations such as the global optimization progress and the remaining optimization budget. The combination of local and global optimization status provides agent comprehensive optimization information about the sub-modules and the tasks. Besides, these features are generic across different problem scopes, which empower ConfigX the generalization ability across unseen problems.

\section{Pseudo Code}\label{appx:pseudo}

In this section to enhance the clarity and overall comprehension of the paper, we present the pseudo code for the RL training process in Algorithm~\ref{alg:pseudo}.

\begin{algorithm}[tb]
\caption{Pseudocode for the overall training.}
\label{alg:pseudo}\footnotesize
\textbf{Input}: The training task set $T_\text{train}$, the ConfigX policy $\pi_\theta$ and critic $V_\phi$, PPO parameters $nstep=10$ and $\kappa=3$ \\
\textbf{Output}: A well-trained policy $\pi_{\theta^*}$.
\begin{algorithmic}[1] 
\For {$epoch \leftarrow 1$ to $100$}
   \For {task $env \in T_\text{train}$}
        \State Initialize the transition memory $MT \leftarrow \emptyset$
        \State Environment initialization $s_1 \leftarrow env.$init() 
        \For {$t \leftarrow 1$ to $H$}
            \State Get the algorithm configuration $a_t \leftarrow \pi_\theta(s_t)$
            \State Execute one optimization step with the given configuration $s_{t+1}, r_t \leftarrow env.$step($a_t$)
            \State $MT \leftarrow MT \bigcup <s_t, a_t, s_{t+1}, r_t>$
            \If {$\mod(t, nstep) = 0$}
                \For {$k \leftarrow 1$ to $\kappa$}
                    \State Update $\theta$ and $\phi$ using $MT$ in PPO manner
                \EndFor
                \State Clear memory $MT \leftarrow \emptyset$
            \EndIf
        \EndFor
    \EndFor
\EndFor
\end{algorithmic}
\end{algorithm}

In the pseudocode we omit the batch processing on tasks for better readability. For each training epoch and each training task form the training task set $T_\text{train}$, we first initialize a memory to contain the transitions. Then we consider the task as a RL environment and initialize it, in which the optimization population is generated and evaluated to obtain the initial optimization state $s_1$ whose formulation is detailed in Appendix~\ref{appx:state}. For each optimization step, the action (configuration) $a_t$ is determined by the policy $\pi_theta$ according to the current state. The algorithm in the environment (task) then takes an optimization step  with the given configuration and returns the next state $s_{t+1}$ and the reward $r_t$. For each 10 optimization steps, the policy parameters are updated for $\kappa=3$ steps in PPO manner using the collected transitions.

\section{Experiment}\label{appx:exp}
\subsection{Experiment Setup}\label{appx:exp_setup}

\subsubsection{Optimization Benchmarks}
In this paper to construct $\mathcal{I}_\text{syn}$ and $\mathcal{I}_\text{real}$ we introduce three optimization problem sets: BBOB testsuite~\cite{bbob2010}, Protein-docking benchmark~\cite{proteinbench} and HPO-B benchmark~\cite{hpo-b}. 
\begin{itemize}
    \item \textbf{$\mathcal{I}_\text{syn}$}. In this problem space we adopt the BBOB testsuite~\cite{bbob2010} which contains 24 optimization problems with diverse properties including unimodal or multi-modal, separable or non-separable, adequate or weak global structure, etc.. In experiments we set the search space to $[-5,5]^D$ and randomly select the dimension $D$ of each problem from $\{5, 10, 20, 50\}$. Besides, for each problem we randomly introduce a offset $z\sim\mathcal{U}(-4, 4)$ to the optimal and then randomly generate a rotational matrix $M\in\mathbb{R}^{D\times D}$ to rotate the searching space. The yielded transformed problems $f(M^\text{T}(x - z))$ construct the problem instance space $\mathcal{I}_\text{syn}$. 
    \item \textbf{$\mathcal{I}_\text{real}$}. For this problem space we combine two realistic problem benchmarks: Protein-docking benchmark~\cite{proteinbench} and HPO-B benchmark~\cite{hpo-b}. 
    \begin{itemize}
        \item \textbf{Protein-docking benchmark}~\cite{proteinbench} aims at minimizing the Gibbs free energy resulting from protein-protein interaction between a given complex and any other conformation. We formulate the objective as follows:
        \begin{equation}
            \min_{x} E_{int}(x,x_0) = \sum_{i}^{atoms} \sum_{j}^{atoms} E(x^i,x_0^j),
        \end{equation}
        where $E(x^i,x_0^j)$ is the energy between any pair atoms of $x$ and $x_0$, and is  defined as :
        \begin{equation}
        \resizebox{0.9\columnwidth}{!}{$
        E_{i,j}= \begin{cases}
        \frac{q_iq_j}{\epsilon r_{i,j}} + \sqrt{\epsilon_i\epsilon_j} 
         \left[(\frac{R_{i,j}}{r_{i,j}})^{12} - (\frac{R_{i,j}}{r_{i,j}})^6\right],  & 
         \text{if }  r_{i,j} < 7 \\ \\
        \left [\frac{(r_{off - f_{i,j}})^2(r_{off} + 2r_{i,j}-3r_{on})}{(r_{off}-r_{on})^3}\right ]\cdot \\\left\{ \frac{q_iq_j}{\epsilon r_{i,j}} + \sqrt{\epsilon_i\epsilon_j}\left[(\frac{R_{i,j}}{r_{i,j}})^{12} - (\frac{R_{i,j}}{r_{i,j}})^6\right] \right\}, & \text{if } 7 \leq r_{i,j} \leq
         9\\ \\
         0  & \text{if } r_{i,j} > 9
        \end{cases}$
        }
        \end{equation}
        All parameters and calculations are taken from the Charmm19 force field~\cite{mackerell1998all}. We select 8 protein-protein complexes from the benchmark and associate each complex with 10 different starting points, chosen from the top-10 start points identified by ZDOCK~\cite{pierce2014zdock}. Consequently, the Protein-Docking testsuites in this paper comprise a total of 80 docking problem instances. It is important to note that we parameterize the search space as $\mathbb{R}^{12}$, which is a reduced dimensionality compared to the original protein complex~\cite{cao2020bayesian}. 
        \item \textbf{HPO-B benchmark}~\cite{hpo-b} includes a wide range of hyper-parameter optimization tasks for 16 different model types (e.g., SVM, XGBoost, etc.). These models have various search spaces ranging from $[0,1]^2$ to $[0,1]^{16}$. Each model is evaluated on several datasets, resulting in a total of 86 tasks. In this paper, we adopt the continuous version of HPO-B, which provides surrogate evaluation functions for time-consuming machine learning tasks to save evaluation time. 
    \end{itemize}
\end{itemize}
\subsubsection{Algorithm Space Split}
We have leveraged the convenience provided by our proposed Modular-BBO to construct two different algorithm structure sub-spaces: $\mathcal{A}_\text{DE}$ and $\mathcal{A}_\text{PSO,GA}$. This algorithm space split help us validate if ConfigX could generalize far beyond the algorithm space it has been trained on. 
\begin{itemize}
    \item \textbf{$\mathcal{A}_\text{DE}$}. In this algorithm structure sub-space, all possible algorithm structures are DE algorithms. To construct such DE algorithm space, we add additional constraints during the algorithm structure generation structure. Concretely, for \textsc{Uncontrollable} sub-modules, they are shared among different EA types hence can be selected without constraints. For \textsc{Controllable} sub-modules, we only constraints the optional range within those related with DE. 
    \item \textbf{$\mathcal{A}_\text{PSO,GA}$}. Same like $\mathcal{A}_\text{DE}$, except that for \textsc{Controllable} sub-modules, we only constraints the optional range within those related with PSO and GA.
\end{itemize}

\subsubsection{Task Set Construction}
We provide multiple zero-shot generalization scenarios in experiments part to validate the zero-shot performance of ConfigX on in-distribution and out-of-distribution tasks. Concretely, we have prepared $4$ task sets for the training and testing of ConfigX:
\begin{itemize}
    \item \textbf{$T_\text{train}$}. A group of $256$ tasks within the joint task space combining the DE algorithm sub-space and synthetic problem space: $T_\text{train}\subset \mathcal{A}_\text{DE} \times \mathcal{I}_\text{syn}$. Concretely, we generate $32$ DE algorithms from $\mathcal{A}_\text{DE}$ and select $8$ problem instances from $\mathcal{I}_\text{syn}$. 
    \item \textbf{$T_\text{test,in}$}. A group of $512$ tasks within the joint task space combining the DE algorithm sub-space and synthetic problem space: $T_\text{test,in}\subset \mathcal{A}_\text{DE} \times \mathcal{I}_\text{syn}$. Concretely, we generate $32$ DE algorithms from $\mathcal{A}_\text{DE}$ and select $16$ problem instances from $\mathcal{I}_\text{syn}$. 
    \item \textbf{$T^{(1)}_\text{test,out}$}. A group of $512$ tasks within the joint task space combining the DE algorithm sub-space and synthetic problem space: $T^{(1)}_\text{test,out} \subset \mathcal{A}_\text{DE} \times \mathcal{I}_\text{real}$. Concretely, we generate $32$ DE algorithms from $\mathcal{A}_\text{DE}$ and select $16$ problem instances from $\mathcal{I}_\text{real}$, 8 from Protein-docking benchmark and 8 from HPO-B. 
    \item \textbf{$T^{(2)}_\text{test,out}$}. A group of $512$ tasks within the joint task space combining the PSO and GA algorithm sub-space and synthetic problem space: $T^{(2)}_\text{test,out} \subset \mathcal{A}_\text{PSO,GA} \times \mathcal{I}_\text{syn}$. Concretely, we generate $32$ PSO and GA algorithms from $\mathcal{A}_\text{PSO,GA}$ and select $16$ problem instances from $\mathcal{I}_\text{syn}$. 
\end{itemize}
 
\subsection{Baselines and Performance Metric}\label{appx:exp_norm}

\subsubsection{Baselines.}
In the experiment section in main paper, we consider three baselines: SMAC3, Original and Random. 
\begin{itemize}
    \item \textbf{SMAC3}. It is a Bayesian Optimization~\cite{BO2016} based hyper-parameter optimization software, we consider it as a baseline of human-crafted state-of-the-art automatic configuration. We in the experiments called the exposed interface \emph{samc.facade.AlgorithmConfigurationFacade()} to determine a single well-performing robust configuration for an algorithm in entire optimization process. This interface combines RandomForest surrogate model, logEI acquisition function and aggressive racing intensification mechanism to provide robust AC performance across many AutoML tasks. In the experiments, we allow SMAC3 to search for optimal configurations for each algorithm structure in $\mathcal{A}_\text{DE}$ and $\mathcal{A}_\text{PSO,GA}$ on the $8$ training problem instances in $T_{\text{train}}$. Then we apply the optimized configurations for the generalization performance evaluation on the testing task sets: $T_\text{test,in}$, $T^{(1)}_\text{test,out}$ and $T^{(2)}_\text{test,out}$. 
    \item \textbf{Original}. It uses the sub-modules' default configurations~(see Table~\ref{tab:control}) for optimizing the target optimization problems in the optimization tasks. We note that Original baseline operates the AC tasks differently with ConfigX. Our ConfigX provide dynamic and flexible AC for each optimization step along the horizon. Original, on the other hand, follows the default configurations from the start to the end.  
    \item \textbf{Random}. It randomly selects a configuration value for each sub-modules in an algorithm structure, in each optimization step. This baseline and the Original baseline aid in demonstrating the learning effectiveness of ConfigX.
\end{itemize}

\subsubsection{Performance Metric.}
The scale of the objective values across tasks can vary, to ensure the fairness of the comparison of baselines across tasks, we conduct a normalization process to restrict the objective value scales of different tasks in the same level.
Considering minimization problem, we first calculate the maximum and minimum objective values in each task across all baselines and all runs: 
\begin{equation}
\begin{split}
    Obj^i_\text{max} = &\mathop{\max}\limits_{b\in B, g\in[1,51]}\{f^{i,*}_{b,g,0}\}\\
    Obj^i_\text{min} = &\mathop{\min}\limits_{b\in B, g\in[1,51]}\{f^{i,*}_{b,g,H}\}
\end{split}
\end{equation}
where $f^{i,*}_{b,g,0}$ and $f^{i,*}_{b,g,H}$ are the best objective values found in the $g$-th run of baseline $b$ at optimization step 0 (initial step) and $H$ (last step) on $i$-th task, $H$ denotes the length of optimization horizon and $B$ is the set of baselines including ConfigX, SMAC3, Original and Random. $Obj^i_\text{max}$ and $Obj^i_\text{min}$ denote the  objective value upper and lower bounds of the $i$-th task. To make the performance metric consistent across all experiments in this paper, we calculate $Obj_\text{max}$ and $Obj_\text{min}$ for all tasks using the experiment data in Section 4.2 and fix them in the performance normalization in all experiments. Then we use these upper and lower bounds of each task to min-max normalize the performance of baseline $b$ at optimization step $t$ as:
\begin{equation}
    Obj_{b, t} = \frac{1}{K\cdot N}\sum_{i=1}^{K\cdot N}\left[ \frac{1}{51}\sum_{g=1}^{51} \frac{f^{i,*}_{b,g,t} - Obj^i_\text{min}}{Obj^i_\text{max} - Obj^i_\text{min}}\right]
\end{equation}
where $K\cdot N$ denotes the number of tasks. The corresponding min-max normalized performance used in our experimental results are computed as $(1-Obj)$. This normalization process is applied on all performance results in the experiments.

\subsection{Additional Experiment Results}\label{appx:exp_add}
\subsubsection{Show Case on Advanced Algorithms}
\begin{figure}[t]
    \centering   
    \includegraphics[width=0.99\columnwidth]{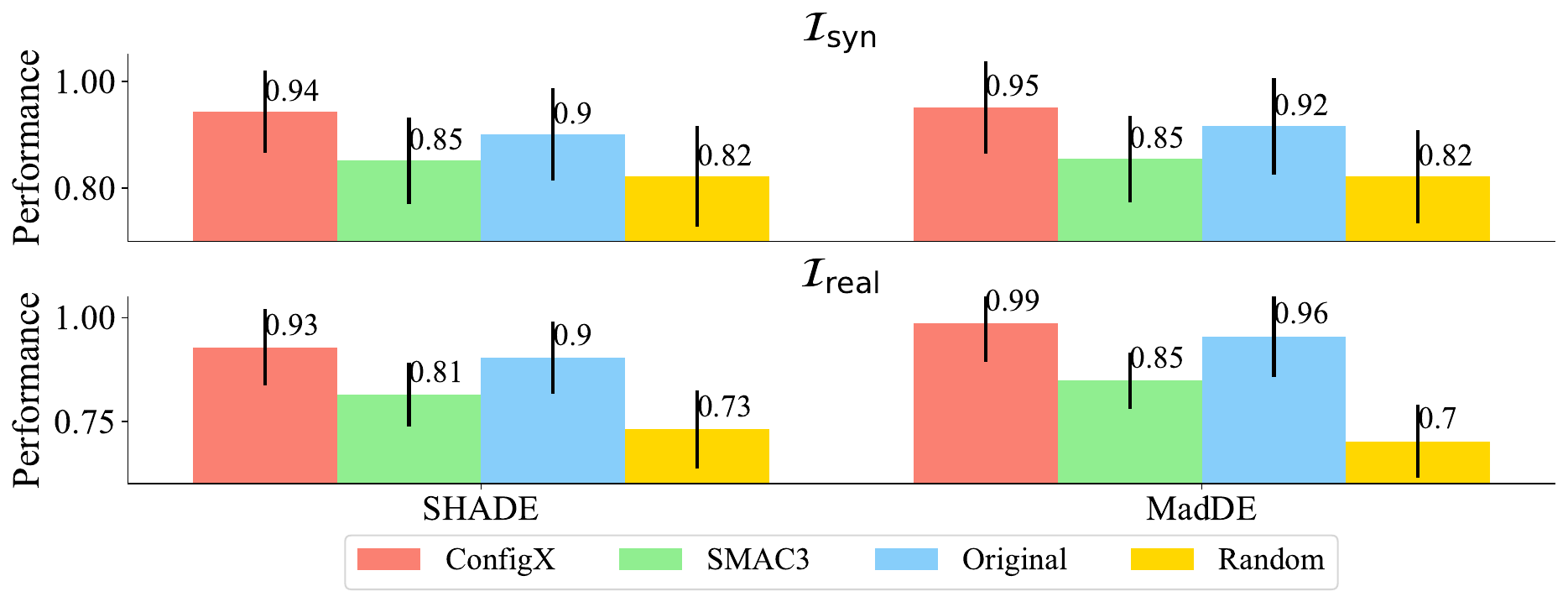}
    \caption{The performance on advanced DE tasks.}
    \label{fig:adv_de_hist}
\end{figure}

One of the advantage in ConfigX against the other MetaBBO methods is that our ConfigX is capable of configuring any algorithm structures within the Modular-BBO's algorithm space. We here demonstrate the practical usage of ConfigX by showcasing its adaptability for practical EAs such as SHADE~\cite{SHADE} and MadDE~\cite{MadDE} which represent the first-rank optimization performance. We directly use the pre-trained ConfigX model to configure them and present the optimization performance of the pre-trained model and other baselines in Figure~\ref{fig:adv_de_hist}. The results show that: (a) Our pre-trained model is sufficient to boost performance of up-to-date EAs. (b) SMAC3 falls short in boosting such EAs since on the one hand, these EAs have intricate configuration spaces hence challenging the searching ability of SMAC3, and on the other hand SMAC3 searches for a good configuration for the whole optimization process hence not as flexible as ConfigX. 

\begin{table}[t]
\centering
\resizebox{\columnwidth}{!}{%
\begin{tabular}{c|cccc}
\hline
         & $T_\text{train}$ & $T_\text{test,in}$ & $T^{(1)}_\text{test,out}$ & $T^{(2)}_\text{test,out}$ \\ \hline
ConfigX  & 527s / 7.03s      & -- / 7.14s     & -- / 8.64s       & -- / 7.11s       \\
SMAC3    & 281s / 5.86s      & 286s / 5.93s     & -- / 7.53s        & 284s / 5.89s       \\
Original & -- / 5.73s      & -- / 5.76s     & -- / 7.49s       & -- / 5.72s       \\
Random   & -- / 5.79s      & -- / 5.80s     & -- / 7.55s       & -- / 5.79s       \\ \hline
\end{tabular}%
}
\caption{The training / testing time efficiency of ConfigX and the baselines on the four task sets.}
\label{tab:time}
\end{table}

\subsubsection{Time Efficiency Comparison}

In this section we investigate the time efficiency of ConfigX and the baselines. For training, we evaluate ConfigX by averaging its training time in 50 epochs and measure SMAC3's per task training time with its default settings (i.e., 100 \textit{n\_trails}). For testing, we evaluate the averaged per task per run testing time. In Table~\ref{tab:time} we present the averaged training / testing time on all four task sets, `--' means no training time required. (a) Although ConfigX requires almost double training time of SMAC3 on $T_\text{train}$, ConfigX can zero-shot to unseen algorithm structures without further training time. SMAC3 requires scratch training for all algorithms and leads to a larger overall training time on all four task sets, which highlights the time efficiency advantage of ConfigX against SMAC3. (b) On the testing time, ConfigX takes more time than the three baselines, due to the network processing. But the superior performance of ConfigX proves the value of the extra runtime.

\subsubsection{Train-Test Split Analysis}

In this section we conduct the experiments on different train-test splits to analyze the impact of different training and test set allocations. Concretely, we train ConfigX models on doubled and halved training tasks set with 64 and 16 DE algorithms generated from $\mathcal{A}_\text{DE}$ respectively, combining with the 8 instances in $\mathcal{I}_\text{syn}$. Then we compare them with the model trained on normal size training task set with 32 DE algorithms on the three testing task sets $T_\text{test,in}$, $T^{(1)}_\text{test,out}$ and $T^{(2)}_\text{test,out}$. The results presented in Table~\ref{tab:traintest-sp} shows that increasing training instances slightly improves generalization, while reducing instances leads to performance degradation. However, considering the doubled training time and reduced training efficiency, the brought performance improvement is limited. Therefore in this paper we use 32 DE algorithms for training which balances the performance and training efficiency.

\begin{table}[t]\small
\centering
\begin{tabular}{c|ccc}
\hline
 &
  $T_\text{test,in}$ &
  $T^{(1)}_\text{test,out}$ &
  $T^{(2)}_\text{test,out}$ 
  \\ \hline
ConfigX     
&  \begin{tabular}[c]{@{}c@{}}9.81E-01\\ $\pm$7.33E-03\end{tabular}
&  \begin{tabular}[c]{@{}c@{}}9.86E-01\\ $\pm$2.64E-03\end{tabular}
&  \textbf{\begin{tabular}[c]{@{}c@{}}9.22E-01\\ $\pm$6.94E-03\end{tabular}}
\\  \hline
ConfigX-Double 
&  \textbf{\begin{tabular}[c]{@{}c@{}}9.83E-01\\ $\pm$6.94E-03\end{tabular}}
&  \textbf{\begin{tabular}[c]{@{}c@{}}9.87E-01\\ $\pm$2.86E-03\end{tabular}}
&  \textbf{\begin{tabular}[c]{@{}c@{}}9.22E-01\\ $\pm$5.28E-03\end{tabular}}
\\ \hline
ConfigX-Half 
&  \begin{tabular}[c]{@{}c@{}}9.76E-01\\ $\pm$8.63E-03\end{tabular}
&  \begin{tabular}[c]{@{}c@{}}9.82E-01\\ $\pm$9.11E-03\end{tabular}
&  \begin{tabular}[c]{@{}c@{}}9.19E-01\\ $\pm$9.01E-03\end{tabular}
\\ \hline
\end{tabular}%
\caption{Performance of ConfigX models trained with different training sets.}
\label{tab:traintest-sp}
\end{table}

\subsubsection{Sub-module Set Size Analysis}

To investigate how might the size of sub-module set impact the model's performance, we construct 
a reduced sub-module set which only contains necessary sub-modules for complete DE algorithms: \textsc{Initialization}, \textsc{Mutation}, \textsc{Crossover}, \textsc{Boundary\_Control} and \textsc{Selection}. Then we generate 32 DE algorithms from the reduced sub-module set and obtain a training task set $T'_\text{train}$ with the 8 instances in $\mathcal{I}_\text{syn}$. The performance of ConfigX and the model trained on $T'_\text{train}$ (denoted as ``ConfigX-Reduce'') is shown in Table~\ref{tab:reduce-mod}. The results show that the narrowed sub-module set does not influence the RL learning effectiveness, while the generalization of the learned model is influenced. The model trained with limited sub-modules fails to apply its configuration experience on unseen sub-modules, which is consistent with our conclusions in Section~\ref{sec:zeroshot}.

\begin{table}[t]\scriptsize
\centering
\begin{tabular}{c|cccc}
\hline
 &
  $T'_\text{train}$ &
  $T_\text{test,in}$ &
  $T^{(1)}_\text{test,out}$ &
  $T^{(2)}_\text{test,out}$ 
  \\ \hline
ConfigX     
&  \begin{tabular}[c]{@{}c@{}}9.84E-01\\ $\pm$6.54E-03\end{tabular}
&  \textbf{\begin{tabular}[c]{@{}c@{}}9.81E-01\\ $\pm$7.33E-03\end{tabular}}
&  \textbf{\begin{tabular}[c]{@{}c@{}}9.86E-01\\ $\pm$2.64E-03\end{tabular}}
&  \textbf{\begin{tabular}[c]{@{}c@{}}9.22E-01\\ $\pm$6.94E-03\end{tabular}}
\\  \hline
ConfigX-Reduce 
&  \textbf{\begin{tabular}[c]{@{}c@{}}9.87E-01\\ $\pm$6.88E-03\end{tabular}}
&  \begin{tabular}[c]{@{}c@{}}9.79E-01\\ $\pm$7.55E-03\end{tabular}
&  \begin{tabular}[c]{@{}c@{}}9.81E-01\\ $\pm$7.69E-03\end{tabular}
&  \begin{tabular}[c]{@{}c@{}}9.18E-01\\ $\pm$8.31E-03\end{tabular}
\\ \hline
\end{tabular}%
\caption{Performance of ConfigX models trained with different sub-module sets.}
\label{tab:reduce-mod}
\end{table}

\bigskip

\bibliography{aaai25}

\end{document}